%% file: main.tex
\documentclass[acmsmall,screen,authorversion]{acmart}

\AtBeginDocument{%
  \providecommand\BibTeX{{%
    \normalfont B\kern-0.5em{\scshape i\kern-0.25em b}\kern-0.8em\TeX}}}

\setcopyright{acmcopyright}
\copyrightyear{2023}
\acmYear{2023}
\acmDOI{10.1145/3596907}

\acmJournal{CSUR}
\acmVolume{55}
\acmNumber{14s}
\acmArticle{334}
\acmMonth{7}

\input{includes}

\begin{document}

\title{Federated Learning for Computationally-Constrained Heterogeneous Devices: A Survey}

\author{Kilian Pfeiffer}
\email{kilian.pfeiffer@kit.edu}
\orcid{0000-0003-3872-0495}
\affiliation{%
  \institution{Karlsruhe Institute of Technology}
  \city{Karlsruhe}
  \country{Germany}
}

\author{Martin Rapp}
\email{martin.rapp@kit.edu}
\orcid{0000-0002-5989-2950}
\affiliation{%
  \institution{Karlsruhe Institute of Technology}
  \city{Karlsruhe}
  \country{Germany}
}

\author{Ramin Khalili}
\email{ramin.khalili@huawei.com}
\orcid{0000-0003-2463-7033}
\affiliation{%
  \institution{Munich Research Center Huawei Technologies}
  \city{Munich}
  \country{Germany}
}

\author{Jörg Henkel}
\email{henkel@kit.edu}
\orcid{0000-0001-9602-2922}
\affiliation{%
  \institution{Karlsruhe Institute of Technology}
  \city{Karlsruhe}
  \country{Germany}
}

\renewcommand{\shortauthors}{Pfeiffer, et al.}

\input{0_abstract}



\maketitle

\input{1_introduction}

\input{2_background_in_FL}

\input{3_heterogenity_in_FL}

\input{4_SOTA}

\input{5_open_problems}

\input{6_conclusion}

\begin{acks}
This work is in parts funded by the Deutsches Bundesministerium für Bildung und Forschung (BMBF, Federal Ministry of Education and Research in Germany).
\end{acks}

\bibliographystyle{ACM-Reference-Format}
\bibliography{bib/bibfile}

\end{document}

%% file: includes.tex
\usepackage[capitalise]{cleveref}
\usepackage{pgfplots}
\usepackage{booktabs}
\usepackage{url}
\usepackage{xcolor,soul,lipsum,xspace}
\soulregister\cref7
\soulregister\cite7
\usepackage[normalem]{ulem}
\usepackage{bm}
\usepackage{amsmath}
\usepackage{graphicx}
\usepackage{pifont}
\newcommand{\cmark}{\ding{51}}%
\newcommand{\xmark}{\ding{55}}%
\newcommand{\etal}{\emph{et al.}\xspace}
\usepackage{tikz}
\usetikzlibrary{positioning, arrows}
\usepackage{pgfplots}
\usepackage[flushleft]{threeparttable}
\usepackage{multirow}
\usetikzlibrary{shapes}
\usepackage{siunitx}
\usepackage{tabularx}
\usepackage{acro}
\DeclareAcronym{FL}{short=FL,long=federated learning,short-indefinite=an}
\DeclareAcronym{IoT}{short=IoT,long=internet of things,short-indefinite=an}
\DeclareAcronym{NN}{short=NN,long=neural network,short-indefinite=an}
\DeclareAcronym{CNN}{short=CNN, long=convolutional neural network}
\DeclareAcronym{fedavg}{short=FedAvg, long=Federated Averaging}
\DeclareAcronym{fedmd}{short=FedMD, long=federated model distillation}
\DeclareAcronym{iid}{short=iid, long=independent and identical distributed}
\DeclareAcronym{ML}{short=ML, long=machine learning}
\DeclareAcronym{kd}{short=KD, long=knowledge distillation}
\DeclareAcronym{sgd}{short=SGD, long=stochastic gradient descent}
\DeclareAcronym{NAS}{short=NAS, long=neural architecture search}

\newcommand{\fedavg}[0]{\ac{fedavg}\xspace}
\newcommand{\fedmd}[0]{\ac{fedmd}\xspace}
\newcommand{\Fedmd}[0]{\Ac{fedmd}\xspace}
\pgfplotsset{compat=1.17} 
\usepackage{makecell}

%% file: 0_abstract.tex
\begin{abstract}
With an increasing number of smart devices like \ac{IoT} devices deployed in the field, offloading
training of \acp{NN} to a central server becomes more and more infeasible. Recent efforts to improve users' privacy have led to on-device learning emerging as an alternative. However, a model trained only on a single device, using only local data, is unlikely to reach a high accuracy.
\Ac{FL} has been introduced as a solution, offering a privacy-preserving trade-off between
communication overhead and model accuracy by sharing knowledge between devices but disclosing the
devices' private data. The applicability and the benefit of applying baseline FL are, however, limited in many
relevant use cases due to the heterogeneity present in such environments. In this
survey, we outline the heterogeneity challenges FL has to overcome to be widely applicable in real-world
applications. We especially focus on the aspect of computation heterogeneity among the participating devices and provide a comprehensive overview of recent works on heterogeneity-aware FL. We discuss two groups: works that adapt the NN architecture and works that approach heterogeneity on a system level, covering \fedavg, distillation, and split learning-based approaches, as well as synchronous and asynchronous aggregation schemes.
\end{abstract}

%% file: 1_introduction.tex
\section{Introduction}
In recent years, a paradigm shift in \ac{ML}  on smart devices, such as \ac{IoT} or smartphones, could be observed. Previously, most deployments of \ac{ML} solutions on such devices were designed to train the \ac{ML} model once at design time in a high-performance cloud~\cite{samie2019cloud}. At run time, the fully-trained model gets deployed on the devices, where only inference tasks are performed. The increasing number of smart devices and recent hardware improvements open the possibility of performing (continuous) on-device learning. This paradigm shift is mainly motivated by privacy and security concerns and motivated by policies like the European Union's~GDPR~\cite{eu_gdpr} or California's~CCPA~\cite{cali}.

On-device learning has several advantages over centralized training.
In the extreme case, when devices independently perform on-device learning, no information about local data samples leaves the device, maintaining users' privacy.
Additionally, on-device learning eliminates communication that comes with uploading collected training samples to a centralized server.
This is particularly relevant with the expected growth rate of \ac{IoT} devices~\cite{manyika2015unlocking}, each equipped with sensors producing massive amounts of data, where the increasing communication burden limits the ability to process all this data centrally.
However, this comes with new challenges.
Firstly, not all deployed devices are capable of doing any training, hence rely on an externally trained model. Further, devices that are capable can only train on their own samples (a tiny subset of a potential centralized dataset), and the resulting models suffer from accuracy losses and weak generalization properties.

\Acf{FL}~\cite{mcmahan2017communicationefficient} is a recently introduced decentralized approach, where training is done in a distributed manner on each device, but devices can still collaborate to share knowledge. 
\Ac{FL} improves privacy~\cite{bonawitz2016practical, wei2020federated} compared to the traditional centralized cloud paradigm.
At the same time, \ac{FL} enables devices to exchange relevant knowledge, improving the models' ability to generalize and overall increasing the accuracy.
There are several techniques for how knowledge can be exchanged.
The most common approach is to exchange \ac{NN} model weights~\cite{mcmahan2017communicationefficient}, but there are other methods, as will be discussed later.

While \ac{FL} systems for smart devices are proposed for a lot of different fields like health-care~\cite{yuan2020federated, ju2020federated, gao2019hhhfl, chen2020fedhealth}, transportation~\cite{liang2019federated, saputra2019energy, ciftler2020federated, posner2021federated}, and robotics~\cite{liu2019lifelong, liu2019federated} using natural language processing, computer vision, and reinforcement learning policies, only a few production use cases like the \emph{Google Keyboard (GBoard)}~\cite{yang2018applied} provide evidence of the success of the \ac{FL} approach.
We argue that real-world applications powered by \ac{FL} are challenging to build because of the heterogeneity present in these environments~\cite{bonawitz2019federated,Li2020,imteaj2020federated}, as almost any real-world system has heterogeneous properties that affect the efficacy of \iac{FL} system.
A key factor of heterogeneity is the devices' different capabilities to perform training of \iac{NN} due to different degrees of \emph{computational resources}\footnote{\cref{sec:heterogenity} gives a detailed overview over the types of computational resources, and the sources and characteristics of heterogeneity in these resources.}.
Training \acp{NN} is computationally expensive due to the high number of trained parameters and its iterative search for a solution. This manifests itself in long training times, ranging up to several weeks for complex tasks. Today's \ac{IoT} devices, smartphones, and embedded systems are still heavily constrained in their training capabilities.

For example, the \emph{PM2.5}~\cite{Chen2017,Chen2018} \ac{IoT} sensor network continuously measures air quality (fine particular matter below a diameter of \SI{2.5}{\micro \metre}) to train a model for anomaly detection.
Several factors affect the devices' computational resources.
The two most important ones in this setting are:
\begin{itemize}
    \item The open-source nature of the project allows for a variety of hardware realizations, hence sensor devices have heterogeneous computational resources.
    \item Sensor devices are deployed in various environments, like indoors, where the devices are continuously powered, or outdoors where energy harvesting is required, limiting energy for training in a heterogeneous manner.
\end{itemize}
These and other sources of heterogeneity need to be considered when designing \iac{FL} system for cooperative learning.
However, research on \ac{FL} on computationally-constrained heterogeneous systems is still in its infancy.

For instance, in \emph{GBoard}, incorporating devices with heterogeneous resources is circumvented by forcing a homogeneous setting and excluding devices that do not fit. The \ac{ML} model is exclusively trained on high-end smartphones that are in an idle state and have at least 2GB of memory. These limitations might play a minor role when having a billion participating devices, as it could be the case in \emph{cross-device} \ac{FL}~\cite{Kairouz2019AdvancesAO}, but in smaller-scale applications (i.e., \emph{horizontal cross-silo}~\cite{Kairouz2019AdvancesAO}), excluding a large number of devices from the training reduces the achievable accuracy and generalization of the model. 
More importantly, there are cross-device cases where excluding devices also excludes an essential share of data that is exclusively available on constrained devices~\cite{maeng2022towards}. Because of fairness or fair representation, it might be required to learn from these devices. Therefore, computation-heterogeneity-aware \ac{FL} is required to enable \ac{FL} to learn from all devices and utilize any data.

While we mainly focus on computation heterogeneity in \ac{FL}, heterogeneity also manifests itself in other domains. Devices have different data distributions and quantities of samples available. Also, they could have different communication capabilities. For a wider use of \ac{FL} systems, these heterogeneities should be taken into account.

\subsection{Scope and Contribution}
This survey studies \ac{FL} under computation heterogeneity.
We also briefly discuss other sources of heterogeneity, such as communication and data heterogeneity, when there is an overlap with computation, but we refer readers to recently published surveys, e.g.,  ~\cite{shi2020communicationefficient} and~\cite{kulkarni2020survey}, for a more detailed discussion.
In this survey, we first provide an extensive analysis of the sources of heterogeneity in devices in various kinds of environments and the implications for cross-device and horizontal cross-silo \ac{FL}.
We then provide a thorough analysis of the state-of-the-art techniques that cope with heterogeneous computation capabilities during \ac{FL} training. We focus on literature that tackles computation heterogeneity on two different levels and exclude techniques that improve the resource efficiency of devices through hardware design considerations, such as accelerators~\cite{armeniakos2022hardware}, as they are not specific to \ac{FL}. For techniques that exclusively target inference, such as federated \ac{NAS} techniques, we refer to Lui~\etal~\cite{liu2021federated}.

The presented techniques are grouped into two major groups, namely techniques that tackle heterogeneity through the devices' \ac{NN} architectures level and techniques that address heterogeneity on the system level. An additional fine-grained categorization is based on the employed \ac{FL} paradigm (\emph{\fedavg}, \emph{distillation}, and \emph{split learning}) for NN architecture-level techniques and is based on whether system-level techniques employ synchronous or asynchronous aggregation. We present our taxonomy of the research on \ac{FL} with computation heterogeneity in \cref{sec:heterogenity} and \cref{tab:techniques}, where we outline the different computation-related challenges that come from real-world applications and present a selection of works that make notable contributions to computation heterogeneity aware \ac{FL}. Finally, we conclude by outlining open problems and remaining challenges.
In contrast to previous surveys~\cite{abdulrahman2020survey, xu2021asynchronous, imteaj2021survey, khan2021federated, lo2021systematic, yin2021comprehensive, bellavista2021decentralised} that cover certain aspects of device heterogeneity in \ac{FL}, we provide the following novel contributions:
\begin{table}[t]
    \caption{Previous surveys in FL that partially cover computation heterogeneity in \ac{FL}.}
    \input{src/survey_table}
    \label{tab:competing_surveys}
\end{table}
\begin{itemize}
    \item We provide an up-to-date review of the state of the art and analyze many techniques that are not yet covered in existing surveys. The problem of computation heterogeneity in \ac{FL} only very recently gained relevance with upcoming training-capable \ac{IoT} devices and proposals for 5G sensor networks. Therefore,  \ac{FL} techniques addressing computation heterogeneity only recently gained popularity. In particular, about $50\%$ of our covered techniques were published in 2021/2022.
    Besides, existing surveys cover less than half of the papers we survey.
    \cref{tab:competing_surveys} presents a comparison to other works.
    \item Existing literature treats computation resources in a very abstract way (e.g., only considering the training time of the number of multiply-accumulate operations), neglecting the different kinds of computational resource limitations and their different implications. This is as their main focus is different (see \cref{tab:competing_surveys}). For the design of future real-world FL applications, these abstract metrics do not suffice, as they do not well reflect the variety of heterogeneity sources that affect a deployed FL application. In contrast, we distinguish between four different concepts, which are constraint types, heterogeneity types, the scale of the heterogeneity, and its granularity.
    \item In contrast to existing surveys, we also include distillation-based \ac{FL} approaches. This novel knowledge aggregation technique potentially enables \ac{NN} model-agnostic \ac{FL}, therefore, enables the use of custom-tailored \ac{NN} models to better account for the devices' heterogeneous capabilities. We provide an in-depth description of how distillation-based \ac{FL} approaches exchange knowledge, present seven different approaches that utilize distillation to cope with computation heterogeneity and discuss their current limitations.
\end{itemize}
The remainder of this survey is structured as follows:
First, in~\cref{sec:aggregation_algorithms}, we introduce the major baseline algorithm of \ac{FL}, \fedavg, and recently introduced distillation and split learning approaches and their respective advantages and disadvantages. In~\cref{sec:heterogenity}, we analyze the different sources that enforce computation constraints on devices and discuss how that leads to computation heterogeneity in \ac{FL}. In \cref{sec:comparison}, state-of-the-art work addressing computation heterogeneity is discussed.
Finally, \cref{sec:open_problem} presents open problems and future directions.

%% file: src/survey_table.tex
\small
\begin{center}
	\begin{tabular}{llcc}
	    \toprule
 		\textbf{Survey}  & \textbf{Focus} & \makecell{\textbf{Techniques w/ comp.}\\ \textbf{ heterogeneity}} & \textbf{Distillation} \\
		\midrule
		Abdulrahman~\etal~\cite{abdulrahman2020survey}   & \ac{FL} Overview & 7 & \xmark\\ 
		
		Xu~\etal~\cite{xu2021asynchronous} & Asynchronous \ac{FL} & 4 & \xmark \\ 
		
		Imteaj~\etal~\cite{imteaj2021survey}  &  \ac{FL} for constrained \ac{IoT} & 10 & \xmark \\ 
		
		Khan~\etal~\cite{khan2021federated}  &  \ac{FL} for \ac{IoT} & 3 & \xmark   \\ 
		
		Lo~\etal~\cite{lo2021systematic} &  \ac{FL} Engineering Aspects & 6 & \xmark \\ 
		
		Yin~\etal~\cite{yin2021comprehensive}   & \ac{FL} Overview & 3 & \xmark\\ 
		
		Bellavista~\etal~\cite{bellavista2021decentralised}  & \ac{FL} Deployment & 7& \xmark \\ \midrule
		
		Ours & Comp. heterogeneity in FL & 35 & \cmark \\
		\bottomrule
	\end{tabular}
\end{center}

%% file: 2_background_in_FL.tex
\section{Basics of Federated Learning}
\label{sec:aggregation_algorithms}

\subsection{Problem Formulation}
Similar to distributed \ac{sgd}, \ac{FL} follows a \emph{server-client} model, where client workers (devices) do training and communicate with a central server to share knowledge.
The simplified case of \ac{FL} aims to learn a model under the constraint that the training data is locally distributed among many devices. Therefore the goal is to minimize the following function by finding optimal \ac{NN} weights~$w$ s.t.
\begin{align}
\label{eq:optimization_goal}
        \min_{w} \:\: f(w) \:\: \text{where} \:\: f(w) := \frac{1}{|\mathcal{K}|} \sum_{k \in \mathcal K} f_k (w),
    \end{align} where $f(w)$ is the global loss (at a centralized server that handles knowledge aggregation) and~$f_k(w)$ is the loss function of device~$k$, where $k$ is a device within the set~$\mathcal{K}$. Each device exclusively has access to its local dataset~$\mathcal{D}_k=\{x_k,y_k\}$ where $x_k$ is the input and $y_k$ is the label. The function~$f_k(w)$, therefore, can be rewritten as
    \begin{align}
        f_k(w) = l(x_k,y_k,w) \quad \{x_k,y_k\}\sim \mathcal{P}\ \forall k.
    \end{align}
    Each device~$k$ draws its samples from the distribution $\mathcal{P}$ resulting in $|\mathcal{K}|$ disjoint splits of the full dataset.
    The accuracy of such a scheme is bounded by the following two bounds. First, a natural upper accuracy bound is the centralized training case where a device has access to the whole dataset.
    The second is a device without any knowledge transfer, only relying on its local data, thus building a natural lower bound to the accuracy. We identify two main goals in computational heterogeneity-aware \ac{FL} techniques: Increasing the convergence speed and reaching a high final accuracy despite having constrained devices.
\subsection{Baseline Federated Averaging (\fedavg)}
\label{subsec:fedavg}
\fedavg is an algorithm for \ac{FL} that was first introduced by McMahan~\etal~\cite{mcmahan2017communicationefficient} and is widely considered a baseline for \ac{FL}.
In the case of \emph{synchronous} \fedavg, training is done in rounds. In each round, every device pulls the current model from the server. Now, each device trains for a fixed amount of mini-batches up to several epochs on its data. After training, each device uploads its models to the server. The server model is updated by averaging all the uploaded models.  In the special case where during the local training phase, each device only applies one gradient step, \fedavg behaves like distributed \ac{sgd}.

\begin{figure}[t!]
    \centering
        \includegraphics[page=1]{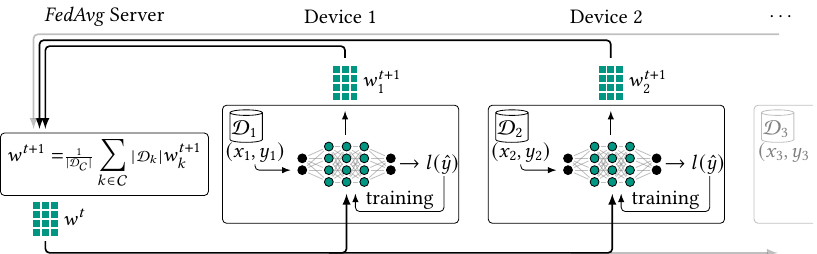}    
    \caption{\fedavg: Knowledge of devices $1,\ldots, K$ is shared through averaging the \acp{NN}' weights. Each device trains on its disjoint local data~$\mathcal{D}_1,\ldots,\mathcal{D}_K$, producing newly trained weights~$w_1,\ldots,w_K$. Every round, the new averaged weights~$w^t$ are distributed to all devices.}
    \label{fig:fed_avg}
\end{figure}

The following detailed description assumes synchronous round-based \fedavg. The aggregation scheme (one round) is visualized in \cref{fig:fed_avg} and described in the following steps:
\begin{enumerate}
    \item \label{item:distribute} At the beginning of every training round, the server deploys the current weights $w^t$ on the set of devices. When starting the training $w^t = w^0$ to achieve the same random initialization of the \ac{NN} on all devices. Since usually a very large number of devices participate in federated learning, a subset $\mathcal C \subset \mathcal{K}$ of all devices is selected for training. 
    \item \label{item:train} Devices train their model (\ac{sgd} steps) for a fixed number of mini-batches or epochs on their local data, consequently, every device does the same amount of training steps
    \begin{align}
        w_k^{t +1} = w^t - \eta \nabla f_k (w^t),
    \end{align} where $w_k^{t+1}$ is the resulting model weights set, while $\eta$ is the learning rate.
    \item Afterward, devices upload their updated model weights $w_k^{t+1}$ to the server.
    \item The server aggregates the devices' knowledge by averaging the received weights using
    \begin{align}
        w^{t+1} = \frac{1}{|\mathcal{D}_{\mathcal C}|}\sum_{k \in \mathcal C } |\mathcal{D}_k| w_k^{t +1},
    \end{align}
    where $w^{t+1}$ represents the new global model and $|\mathcal{D}_{\mathcal C}|$ the number of samples of devices in subset $\mathcal C$. \emph{The next round starts with \cref{item:distribute} ($w^{t+1} \rightarrow w^{t}$)}.
\end{enumerate}
\subsection{Client Selection in \fedavg}
In~\cite{mcmahan2017communicationefficient}, as presented in \cref{subsec:fedavg}, the availability of devices in~$\mathcal{K}$ is modeled to be random, s.t. each round~$t$ a random subset~$\mathcal{C}^t \subset \mathcal{K}$ participates in \ac{FL}. If a device~$k$ in the set~$\mathcal C^t$ takes longer than others, it delays the synchronous aggregation, hence, slows down the overall \ac{FL} training. If waiting becomes impractical, the device has to be dropped from the current \ac{FL} round, wasting its resources. Devices like these are called \emph{stragglers}. Client selection techniques assume that device availability and their expected resources for training can be acquired by the server to select an optimal subset $C^t$ of devices for each round. Further, they allow for variable per-round training time deadlines. Client selection techniques mainly target one or more of the following goals: Maximizing the overall \ac{FL} convergence speed, minimizing the number of stragglers, or minimizing the overall energy spent on~\ac{FL}.

\subsection{Asynchronous \fedavg}
Alternatively, in \emph{asynchronous} \fedavg, devices can pull the most recent model from the server at any time, perform local update steps on it, and upload it at any time. Knowledge aggregation is done at the server as soon as a new model update from a device arrives. \emph{Stale devices}: In asynchronous schemes, devices can not become stragglers since their updated model can be aggregated instantly into the global model. Yet, devices that take too long to finish their training become stale devices. Stale devices upload their trained weights based on an old state of the global model, introducing instability. Chen~\etal first provided evidence of that for asynchronous \ac{sgd}~\cite{chen2017revisiting}. Further, Xi~\etal, as well as Xu~\etal, demonstrate that staleness in \ac{FL} lowers the convergence speed and affects the maximum reachable accuracy~\cite{xie2020asynchronous, xu2019elfish}.

\subsection{Distillation for Federated Learning}
\label{subsec:fedmd}
Distillation techniques in \ac{FL} take motivation from \ac{kd}~\cite{hinton2015distilling}, which was originally used to transfer knowledge of a larger \ac{NN} into a reduced smaller one for model compression. Typically, \acp{NN} trained on classification problems output probabilities using a softmax layer that converts logits into probabilities. \Ac{kd} aims to distill the better generalization of larger models into smaller ones by not training the smaller network on the sample's class but rather on the sample's distribution (\emph{soft label}) that is predicted by the larger network.  Therefore, the smaller network not only learns the correct classes but also, through likelihood scores, learns about the larger model's knowledge representation.

\Fedmd is an algorithm that uses distillation for \ac{FL} that has been proposed recently by Li and Wang~\cite{Li2019}. We exemplary describe how distillation is used in federated learning using \fedmd:
Additional to the devices' local (private) datasets, a second dataset $\mathcal{D}_p$ is introduced, which is a public dataset that is known to all participating devices. With consecutive training of private data and inference on public labels, devices transfer the knowledge of their private data into the public \emph{soft labels.} The major difference to the previously discussed \fedavg scheme is that knowledge is not shared through model weights but through soft labels of a public dataset. This concept change allows for devices to be independent of the server model architecture.

\begin{figure}[t!]
    \centering
        \includegraphics[page=1]{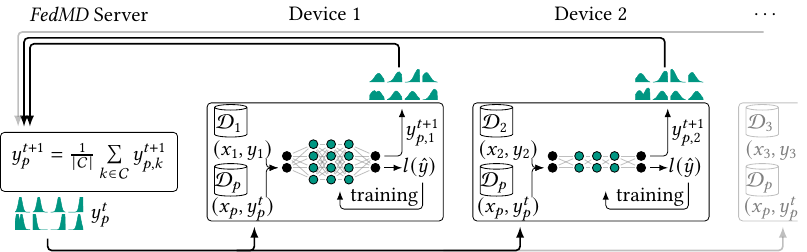}   
    \caption{\fedmd: Knowledge of devices~$1$ to~$K$ is shared through soft labels~$y_p^t$ of the public dataset~$\mathcal{D}_p$ that is known to all devices. Additionally, devices train on their private data set~$\mathcal D_1,\ldots,\mathcal D_K$. Devices transfer knowledge of their private data into public soft labels. Every round, the new averaged soft labels are distributed to the devices.}
    \label{fig:fed_md}
\end{figure}

The steps of a synchronous knowledge aggregation round are visualized in \cref{fig:fed_md} and described in the following steps. In the general case, \fedmd is applicable to any machine learning algorithm and task. For the sake of simplicity, we consider in the following detailed explanation only \acp{NN} that perform classification with a softmax activation layer. Again a subset~$\mathcal{C}$ of all devices contributes in one round.
\begin{enumerate}
    \item \label{item:fedmd_distribute} At the beginning of each round, all devices download the current public dataset's soft labels~$y_p^t$. In the first round, the public dataset's labels are one-hot encoded.
    \item First, all devices train their \acp{NN} on the public dataset~$\mathcal{D}_p$, followed by training on their private data~$\mathcal{D}_k=\{x_k,y_k\}$. After training on the private dataset, an inference phase on the public set follows. Instead of storing the one-hot encoded outputs, each device stores its generated soft label outputs~$y_{p,k}^{t+1}$.
    
    \item The outputs, representing the probability distribution of the input over all possible classes, are uploaded for each public data sample to the server.
    
    \item \label{step:round_md}The results from all participating devices are aggregated by using averaging
    \begin{align}
        { y}_p^{t+1}  = \frac{1}{|\mathcal{C}|} \sum_{k \in \mathcal{C}} { y}_{p, k}^{t+1} ,
    \end{align} building a new averaged soft label for each sample in the public dataset. The public dataset is updated to~$\{x_p,y_p^{t+1}\}$. \emph{The round repeats with \cref{item:fedmd_distribute}~$(y_p^{t+1} \rightarrow y_p^{t})$}.
\end{enumerate}

\subsection{Split Learning}
Lastly, we also briefly elaborate on \emph{split learning} techniques. In difference to \fedavg or distillation-based approaches, split learning techniques transfer information by using activations and gradients. In split learning, the \ac{NN} model is split into two parts: A device model~$a_k = f_k(w_k, x_k)$ for all devices~$k \in \mathcal{K}$, and a single server model~$f_s(w_s, a_k)$ that takes activations~$a_k$ from the devices as input. A combination of both can be used on device~$k$ for inference, s.t.~$\hat y_k = f_s(w_s, f_k(w_k, x_k))$. Server and device models are trained in the following way:
\begin{enumerate}
    \item\label{item:split_learning} Device~$k$ calculates activations~$a_k$ by using~$a_k = f_k(w_k, x_k)$ for its private data~$\mathcal{D}_k$ and sends activations~$a_k$ with the respective class labels~$y_k$ to the server.
    \item The server receives the activations and class labels and calculates~$\hat y= f_s(w_s, a_k)$ as well as the gradient~$\nabla f_s(w_s)$. The gradient step (w.r.t~$f_s$) is applied on the server the gradient w.r.t to~$a_k$ is sent back to device~$k$.
    \item Device~$k$ uses the received gradient from the server to calculate~$\nabla f_k(w_k)$. \emph{The procedure repeats with \cref{item:split_learning}.}
\end{enumerate}
Split learning techniques can reach a high convergence speed and allow to reduce the computational burden on the devices. This comes at the cost of a high communication volume.

%% file: 3_heterogenity_in_FL.tex
\section{Computation Heterogeneity in Federated Learning}
\label{sec:heterogenity}
This section covers different types of heterogeneity in devices and their effects on \ac{FL} systems.

Our main focus is on \emph{computation
heterogeneity in existing devices}, i.e., different devices cooperating in \iac{FL} system differ in their capabilities to train \acp{NN}.

\subsection{Computational Resources in End-Devices}
\label{subsec:computation_heterogenity}

\begin{table}[t]
    \centering
    \caption{(a)~Computational capabilities (float performance, RAM) vary across different device classes. (b)~The resource requirements (number of MAC operations, memory footprint) for training of several established image classification \acp{NN} (training with \emph{PyTorch}~\cite{PyTorch}, on \emph{CIFAR10}~\cite{krizhevsky2009learning} with batch size 32).}
    \setlength{\tabcolsep}{4pt}  
    \setlength{\fboxsep}{0pt}
    \begin{minipage}{0.60\linewidth}
        \centering
        \small
        (a) Computational resources of end-devices
        \input{src/compute_heterogenity_table.tex}
    \end{minipage}
    \begin{minipage}{0.38\linewidth}
        \centering
        \small
        (b) Resource requirements of NNs
        \input{src/table_memory_training.tex}
    \end{minipage}
    \label{tab:compute_heterogenity}
\end{table}

\cref{tab:compute_heterogenity}a shows the computational resources of several common end device classes in terms of floating-point operations per second (FLOPS) and RAM.
These devices range from ultra-low-power MSP430 microcontrollers to high-performance server GPUs.
\cref{tab:compute_heterogenity}b shows the resource requirements of several well-known image classification \acp{NN}, in terms of the number of MAC operations in the forward pass and required memory for training a mini-batch of size 32.
Note that the number of MAC operations for training a whole epoch would be 3-5 orders of magnitude larger because training additionally requires a backward pass, which has around 2${\times}$ the MACs of the forward pass~\cite{ai_and_compute} (plus eventual computations for a stateful optimizer), and a device has hundreds to thousands of local training samples to process in a round.

We observe that it is unrealistic to aim at training recent \ac{NN} topologies on all devices.
For instance, an MSP430-based embedded device by far does not provide sufficient memory.
This is an important observation that puts a limit on how far into the embedded domain we can push \ac{FL}, and it is much more realistic to employ powerful embedded devices such as Raspberry Pi, NVIDIA Jetson, or smartphones in \ac{FL} systems.
Nevertheless, there is large heterogeneity even within one type of devices.
For instance, the computational performance of smartphones varies between $10^{10}$ and $10^{12}$ FLOPS and 512\,MB and 8\,GB RAM.
Heterogeneity greatly affects \ac{FL} because, clearly, not all devices can perform the same computations in each round, as will be discussed in more detail in \cref{subsec:computation_heterogenity_categorization}.

In general, this heterogeneity may have many different sources:
\begin{itemize}
    \item Different hardware or software generations of devices or device tiers may cooperate in one \ac{FL} system. These can be, for instance, different smartphone generations or hardware revisions of \ac{IoT} devices~\cite{Chen2017}. Different hardware generations may be equipped with different \ac{ML} accelerators that speed up training. Devices also can have different amounts of memory or storage capacities, limiting the devices' training capabilities.
    \item Degradation of components affects the available resources. The two famous examples are battery fading, where the energy capacity and peak power capabilities of a rechargeable battery reduce over time~\cite{erdinc2009dynamic,dubarry2007capacity}, and circuit degradation, which reduces the switching speed of a circuit over time, reducing the achievable performance~\cite{van2017reliability}.
    \item The power/energy supply may be subject to variation. For example, the power supplied by a solar cell varies with solar irradiation, which both depends on local random weather effects (clouds) and regular cycles (day/night, summer/winter). Similar effects apply to other energy harvesting techniques~\cite{garg2017energyharvest}.
    \item Ambient temperature affects the efficacy of cooling. This limits the thermally-safe power dissipation and, thereby, also limits computation~\cite{henkel2019smart}.
    \item Shared resource contention with other applications running on a device affects the available resources for training. \Ac{ML} model training often runs in parallel to other tasks on the same platform~\cite{Fang_2018}. This leads to fast-changing degrees of contention in shared resources such as CPU time, memory, or energy.
\end{itemize}

\subsection{Categorization of Constraints and Heterogeneity}
\label{subsec:computation_heterogenity_categorization}
We distinguish between two main categories of computation constraints, \emph{hard constraints} and \emph{soft constraints}, that cause different kinds of computation heterogeneity in an \ac{FL} system, namely, \emph{heterogeneity across devices}, \emph{over rounds}, and \emph{over time}. These heterogeneities can have different \emph{scales} and \emph{granularities}.

\textbf{Hard Constraints}: These constraints prevent a device from training a given \ac{NN} model. The most prominent example is limited memory. Despite considerate efforts to shrink the number of parameters in modern \acp{NN}~\cite{mobilenet, efficientnet}, model architectures like \emph{MobileNetV2}~\cite{howard2017mobilenets} still have millions of parameters, which need to be kept in memory in high-precision (e.g., 32-bit floating-point) during training. In addition to the model parameters, also activations have to be kept in memory for applying backpropagation. This may easily accumulate to more than \SI{1}{\giga\byte} of memory for training, as shown in \cref{tab:compute_heterogenity}b.
If a chosen \ac{NN} architecture in \iac{FL} system exceeds the devices' memory capacity or the memory is not available due to resource contention, the device can not participate in the \ac{FL} system.

\textbf{Soft Constraints}: These constraints allow for training a certain \ac{NN} architecture but prevent the device from achieving its full training throughput (e.g., FLOPS). The computational capability is affected by several factors, such as the used micro-architecture, degradation of components, unstable power supplies, or shared resource contention.
Constraints like these enforce slower training. For a device participating in an \ac{FL} system, soft constraints can prevent the device from finishing its local training on time, making it a straggler or a stale device.

The aforementioned constraints may be heterogeneous throughout the set of devices participating in \ac{FL} and over the training duration. We differentiate between three types of heterogeneity caused by device constraints.

\textbf{Heterogeneity across Devices}: Firstly, devices participating in \iac{FL} system may have different kinds of hard and soft constraints, causing heterogeneity across the devices. These kinds of constraints (e.g., availability of accelerator or memory capacity) are either known at design time or before starting the training and do not change over time. Different devices may be subject to different constraints that limit the training throughput. An example of heterogeneity across devices is a smartphone application \ac{FL} system.
As discussed above, a low-end smartphone operates only with 1/100th of the peak performance and may have only 1/8th of the memory capacity of a high-end smartphone.
\cref{tab:compute_heterogenity} also lists other devices and their respective training throughput and memory resources.
To support heterogeneity across devices, \iac{FL} algorithm needs to be able to use different amounts of resources on different devices in a round.

\textbf{Heterogeneity over Rounds}:
If the number of devices in an \ac{FL} system is comparatively small (cross-silo \ac{FL}), devices need to participate in many \ac{FL} rounds.
This is, for instance, the case in robotics~\cite{imteaj2021fedar} or lifelong learning \cite{liu2019lifelong}.
Soft constraints of devices may be determined by the devices' environments and change over the rounds. Examples of such constraints are the devices' expected battery level during training, the current power supply, or ambient temperatures. This includes all constraints that change slower than FL rounds (in the range of minutes to hours), can be predicted, and are known prior to an FL round.
To support heterogeneity over rounds, \iac{FL} algorithm needs to either implement stateless clients or explicitly support changes in the availability of the resources of a device.

\textbf{Heterogeneity over Time}: Soft constraints that change at a finer granularity than FL rounds cause \emph{heterogeneity over time} in an FL system. These throughput changes can not be predicted and randomly occur (in the range of milliseconds/seconds). One example is resource contention on a smartphone, where the share of available resources is unpredictable for the FL system and changes within seconds, i.e., much faster than FL rounds.
To support heterogeneity over time, \iac{FL} algorithm needs to adaptively adjust the required resources at the client during the round without relying on the server.

In any real system, a combination of the different types of heterogeneities is expected to occur.
Finally, there are two additional properties of the heterogeneity that describe the statistical distribution of constraints present in a set of devices.

\textbf{Scale of Heterogeneity}: 
Heterogeneity across devices or over rounds/time can have different \emph{scales}. This scale can vary depending on the source of constraints. In a smartphone application with different tiers or hardware generations, memory capacity and peak performance can vary by a factor of $10\times$ and $8\times$. Contrary to that, constraints caused by an aging effect (battery) may result in a peak power reduction of more than 50\,\%~\cite{dubarry2007capacity}, which translates to a throughput difference of around 20\%, assuming a cubic relationship between dynamic power and performance ($V^2 f$-scaling~\cite{henkel2015new}).
This is much lower than the scale of $10\times$ observed with different tiers or hardware generations.
\Iac{FL} algorithm must support the scale of heterogeneity present in the system.

\textbf{Granularity of Heterogeneity}: The heterogeneity present in \iac{FL} system can have different \emph{granularities}. In a smartphone application, where devices are equipped with different memory capacities (e.g., $\SI{1}{\giga\byte}$, $\SI{2}{\giga\byte}$, and $\SI{4}{\giga\byte}$), \iac{FL} system has to account for only a small finite number of types of devices. On the other hand, devices experiencing resource contention can have a continuous range of total training throughput in \iac{FL} round.
Either \iac{FL} algorithm supports arbitrary resource availability levels or quantizing the continuous range into a reasonable finite number of constraint levels is required.
The number of supported levels by the algorithm must be high enough to avoid wasting too many resources, which can slow down the overall training.
For example, resource contention (e.g., four other applications sharing CPU time) would reduce the available resources for training by 5$\times$.
If the \ac{FL} algorithm supports only 5 levels, which is not uncommon, as we observe in \cref{sec:comparison}, the ratio between subsequent levels is at least 1.5$\times$ if
levels are distributed exponentially. As an example, levels of $[1\times, 1.5\times,2.2\times,3.3\times,5\times]$ could lead to 33\% of available resources for training being wasted.
Note that this gap increases strongly if a larger scale needs to be supported.

As with all the other properties, \iac{FL} technique must be able to cope with the granularity of the system at hand.

These different types of computational heterogeneity require different solutions.
We analyze the capabilities of the state-of-the-art techniques for heterogeneity-aware FL to cope with all the different types of computational heterogeneity in \cref{sec:comparison}.

\subsection{Communication Heterogeneity}
Knowledge transfer, for instance, through sharing \ac{NN} parameters between devices, is only possible via communication. The throughput, latency, and reliability of communication channels are limited and can vary between devices causing stragglers or stale devices. Most of the current research focuses on making the transmission more efficient by using compression and quantization schemes~\cite{Amiri2020, Mills2020, Konecny2016, elkordy2020secure}. Even though we do not focus on communication heterogeneity in this survey,
we nevertheless cover certain aspects of communication because of the inter-dependencies between communication and computation:
\begin{itemize}
    \item The complexity of the trained \ac{NN} architecture correlates with the model size that has to be transmitted to the server. Hence, reducing the \ac{NN} structure (e.g., by pruning~\cite{jiang2019model}) reduces not only the computation complexity but also the communication burden.
    \item Increasing training throughput and communication throughput can create a trade-off scenario~\cite{pmlr-v108-reisizadeh20a}, because ultimately, both may compete for resources like energy~\cite{mo2020energyefficient}.
\end{itemize}

\subsection{Data Quantity Heterogeneity}
In a real-world scenario, the quantity of the data gathered on different devices may vary. However, to guarantee a high accuracy, the model needs to be trained over all available data. This imposes more training overhead on devices with a higher quantity of data, as they require to perform higher numbers of mini-batch training per round. Techniques that cope with limited throughput (soft constraints) can be applied to such devices to prevent them from becoming a straggler or stale devices.

%% file: src/compute_heterogenity_table.tex
\begin{tabular}{l  r  r}
    \toprule
	\textbf{Device} & \textbf{FLOPS} & \textbf{RAM} \\
	\midrule
    {MSP430 Ser.} & $10^5$ - $10^6$ & $\SI{0.5}{\kilo\byte}$ - $\SI{66}{\kilo\byte}$ \\
    {STM32F7 Ser. (Arm Cortex-M7)} & $2\cdot10^8$ & $\SI{256}{\kilo\byte}$ - $\SI{512}{\mega\byte}$ \\
	{Raspberry Pi Ser.} & $10^8$ - $10^{10}$ & $\SI{512}{\mega\byte}$ - $\SI{8}{\giga\byte}$ \\
	{Low-End Smartphones} & $10^{10}$ - $10^{11}$ & $\SI{1}{\giga\byte}$ - $\SI{2}{\giga\byte}$ \\
	{Nvidia Jetson Nano} & $10^{11}$ - $10^{12}$ & $\SI{2}{\giga\byte}$ - $\SI{4}{\giga\byte}$ \\
	{High-End Smartphones} & $10^{11}$ - $10^{12}$ &  $\SI{4}{\giga\byte}$ - $\SI{8}{\giga\byte}$ \\
	{Server GPUs} & $10^{13}$ - $10^{14}$ & $\SI{32}{\giga\byte}$ - $\SI{100}{\giga\byte}$ \\
	\bottomrule
\end{tabular}

%% file: src/table_memory_training.tex
\begin{tabular}{lcc}
    \toprule
    \textbf{ML Model} & \textbf{\# MACs} & \textbf{Memory} \\
    & {(Forward)} & {(Training)} \\
    \midrule
    {LeNet} & $6.7{\cdot}10^5$ & $\SI{0.5}{\giga\byte}$ \\ 
    {ResNet18} & $5.6{\cdot}10^8$ & $\SI{0.8}{\giga\byte}$\\
    {EfficientNet} & $3.2{\cdot}10^7$ &  $\SI{0.9}{\giga\byte}$ \\
    {MobileNetV2} & $9.6{\cdot}10^7$ & $\SI{1.4}{\giga\byte}$ \\
    {ResNet152} & $3.7{\cdot}10^9$ & $\SI{5.3}{\giga\byte}$ \\
    \bottomrule
\end{tabular}

%% file: 4_SOTA.tex
\section{Computation Heterogeneity-Aware Federated Learning}
\label{sec:comparison}
\subsection{Categorization of Techniques}
We categorize techniques addressing \ac{FL} with heterogeneous computational capabilities into two groups:

\begin{table*}[b]
    \renewcommand{\arraystretch}{1.073}
	\caption{Comparison of techniques aiming at computation heterogeneity-aware \ac{FL}. 
	We differentiate between hard and soft constraints, and heterogeneity across devices (\textbf{D}), over rounds (\textbf{R}), and over time (\textbf{T}).
	}
	    \input{src/taxonomy}
		\label{tab:techniques}
\end{table*}

\ac{NN} architecture level:
We distinguish between techniques where devices can choose from a limited set of model architectures or submodels. These techniques stem from \fedavg and are covered in~\cref{subsec:heterogeneity_aware_subsets}. Other techniques do not impose any limitations on the model architecture. These techniques build on top of \fedmd and are covered in~\cref{subsec:heterogeneity_aware_distillation}. Lastly, we cover split learning based techniques in \cref{subsec:heterogeneity_aware_other}.

System level: In this case, heterogeneity is not addressed by varying model complexity but by allowing for variable-length rounds, grouping, and partial updates. Besides synchronous solutions, variable training times can also be accounted for by allowing for asynchronous updates. These approaches are covered in \cref{subsec:heterogeneity_aware_client_selection,subsec:heterogeneity_aware_async}, respectively.

Lastly, we discuss which types of computation heterogeneity are addressed.  All discussed techniques are listed in \cref{tab:techniques}.

\subsection{\ac{NN} Architecture Heterogeneity based on \emph{FedAvg}}
\label{subsec:heterogeneity_aware_subsets}

The following techniques adapt \fedavg to achieve model architecture heterogeneity. Allowing for variable model complexity in \fedavg is not straightforward since the aggregation scheme relies on averaging of model weights. If the model architectures vary, it is not clear how to match the parameters for averaging. Even networks with the same architecture can have different learned structures, thus, averaging their weights hurts performance~\cite{mcmahan2017communicationefficient}. One reason for that is the \acp{NN}' permutation invariance, which means that even two-layer networks trained on the same distribution can have different internal structures. Several research works, therefore, focus on matching internal features. Wang~\etal~\cite{wang2020federated} introduce \emph{FedMA} to match layer-wise filters together. Similarly, Yurochkin~\etal~\cite{pmlr-v97-yurochkin19a} focus on matching neurons by identifying similar neuron subsets to match features in non-\ac{iid} data scenarios. Reliable feature matching would allow for combining networks with varying architectures as well as a better generalization in non-\ac{iid} data scenarios.
Current techniques circumvent direct feature matching of varying \ac{NN} structures by training with regularly changing variable-sized dropout masks or by training subsets of the full network, as discussed in the following. These techniques share some similarities with well-known inference approaches like pruning. The major differences are that pruning is primarily done after training has converged, intending to create an efficient model for inference. Meanwhile, in the following approaches, smaller subsets remain embedded in the full-size \ac{NN} architecture, while each part is constantly updated every round.

\textbf{Unstructured Subsets:}
Caldas~\etal~\cite{caldas2018expanding} were within the first to address the high computation burden of \ac{FL}, introducing \emph{Federated Dropout} (\emph{FD}) for \fedavg, where instead of the full network, only a subset of the network is trained and updated in every round. A fixed set of weights is set to zero, and the remaining weights are packed into a dense matrix for efficient computation. For convolutional layers, full filters are dropped. In their experiments, they achieve a 1.7$\times$ reduction of computations on \emph{MNIST}~\cite{deng2012mnist} without hurting final accuracy. The provided results suggest that averaging subsets through dropout masks does not negatively impact the aggregation mechanism of \fedavg. While reducing the devices' computation effort, FD forces a fixed dropout rate on all devices, thus limiting the ability to adapt to heterogeneity across devices. Heterogeneity across devices is tackled by Xu~\etal~\cite{xu2019elfish} in \emph{ELFISH}, a method that identifies neurons that contribute much to convergence and builds dropout masks based on this information. Each device receives a specific dropout mask to best match its current computing capabilities. Masks are updated every round. In \emph{DISTREAL}~\cite{rapp2021distreal}, the authors explore how subsets can be trained in environments with time-varying computational resources that change faster than FL rounds and are not known in advance. A mini-batch level granularity for training subsets by randomly switching filters of the CNN during training is achieved. Additionally, DISTREAL does not require a common fixed subset ratio per layer. Instead, this design space is explored with a genetic algorithm to find Pareto-optimal per-layer subset ratios. The results show that in scenarios with fast-changing resources, randomly switching between filters and optimized per-layer subset ratios result in faster convergence and higher final accuracies compared to FD.

\textbf{Structured Subsets:}
While FD and ELFISH utilize unstructured subsets (masks), where the subset parameters are scattered over the full-size \ac{NN} structure, some other studies propose a strictly hierarchical nesting of subsets.
\emph{HeteroFL} is presented by Diao~\etal~\cite{diao2020heterofl}, a \fedavg adaption that allows devices to select from specific subsets of the full model. A smaller set is constructed as a subset of the next bigger set, giving devices a hierarchical selection of networks. Aggregation is done by only averaging trained parameters from the devices. Therefore, some parts of the model are only updated by strong devices. Similarly, Yu and Li~\cite{Yu2021Federated} propose partitioning of \ac{CNN} layer width, depth, and kernel size by slices of power of two and introduce a submodel search algorithm to best match a submodel to the devices' individual resources. They only provide proof-of-concept experiments for small networks in homogeneous cases. Horv\'{a}th~\etal\cite{horvath2021fjord} present \emph{FjORD}, where devices receive submodels of various complexities through applying \emph{Ordered Dropout (OD)}. Differently to HeteroFL, in each local round, every device uniformly selects from different complexity levels (within its capabilities) for a short training period. Since higher-performance devices are not fully utilized this way, the authors apply \ac{kd} on top of OD to transfer knowledge from larger complexity models to smaller ones during local training. For comparison, they extend FD with variable dropout rates for each device and show that their structured subsets outperform random dropout masks.

\textbf{Hybrid Subsets:}
Also, a mixture of both approaches, specifically the use of structured but not hierarchical subsets, is considered. In difference to the previous approaches, this allows training of \ac{NN}-models that exceed the strongest devices' capacity. Additionally, each parameter gets eventually trained by each device. The use of a rolling window approach is proposed by~Alam~\etal~\cite{alam2022fedrolex} in \emph{FedRolex}. In each training round, a device trains a different slice of the \ac{NN}. Federated \ac{NAS}~\cite{he2022fednas,mushtaq2021spider,yao2021federated} techniques are proposed, using subsets of shared common structure to allow for device personalizing, aiming for better accuracy in non-\ac{iid} cases and efficient models for inference. An advantage compared to previous techniques is that the devices' \ac{NN} models can be independently optimized for their hardware, however, this comes at the cost of exploring the architecture search space, which can be resource-hungry. Dudziak~\etal~\cite{dudziak2022fedoras} present \emph{FedorAS}, a federated \ac{NAS} technique that also accounts for device heterogeneity during training. Similarly to FedRolex, devices receive a subset of the full server model, and depending on their resources, switch between further splits of the subset on a mini-batch level. The common training is followed by an architecture search for several tiers based on the full model and lastly, a fine-tuning step.

\textbf{Low-Rank Factorization:}
Low-rank factorization is considered to select the subset's parameters. These techniques also root from inference compression. The major difference to its use for inference is that here, the low-rank \ac{NN} is updated during training, and the low-rank updates are applied to the full model on the server. Yao~\etal\cite{yao2021fedhm} present \emph{FedHM}, where they create low-complexity submodels on the server by doing a low-rank factorization of the full model. Layer parameters with dimensions $m \times n$ are decomposed into two matrices with dimensions $m \times r$ and $r \times n$. The complexity of the model can be controlled via the rank $r$. Computationally weak devices perform two lightweight convolution operations based on the matrix decomposition instead of one complex operation.  To avoid a strong accuracy degradation, the complexity reduction through matrix decomposition is preferably applied in the later layers of the \ac{NN} model. The authors show that subsets through low-rank factorization can achieve higher accuracies than straightforward splitting, but more importantly, dramatically reduces the communication burden. A similar technique is employed by Mei~\etal~\cite{mei2022resource} in \emph{FLANC}, where in the factorization, the parameters are decomposed in matrices with dimensions smaller than $m$ and $n$, allowing for a reduction of the activation size and hence, memory requirements.

\textbf{Others:}
In \emph{ZeroFL}~\cite{qiu2022zerofl}, dropout masks in combination with sparse convolutions are used to lower the computational complexity in training (FLOPS) and reduce the communication volume, although special hardware and software support is required to enable real-world gains.
Lastly, in \emph{CoCoFL}~\cite{pfeiffer2022coco}, a technique is presented that does not use subsets of \iac{NN} for training. Instead, only for some layers per round gradients get calculated while the remainder of the layers are frozen. This saves computation time (fewer backpropagation components) and reduces the upload volume since only updated layers must be uploaded. Further, the freezing of the remaining layers allows for the reduction of the computation complexity by using inference techniques, such as batch norm folding and quantization (e.g., \texttt{int8} instead of \texttt{float32}). While in width scaling approaches (HeteroFL), gains in computation time and communication volume are tightly coupled, selective freezing and training can decouple those properties. The results show benefits over HeteroFL, especially in scenarios where the devices' constraints regarding computation and communication are decoupled.

\textbf{Discussion:}
Most presented techniques send lower complexity subsets of the full NN model to the devices. Allowing flexible NN models addresses hard constraints (e.g., devices can train a submodel with a lower memory footprint). Approaches partly require the resources to be known prior to the round, limiting real-world use cases. A remaining challenge is the scale and granularity of heterogeneity: For example, HeteroFL uses $5$ subsets and scales down the parameters exponentially down to a $250\times$ reduction. Consequently, there is a $4\times$ gap between the full model and the largest sub-model. It remains unclear if subsets maintain effectiveness under both large scale and high granularity (many subsets). Lastly, while devices are enabled to participate in training with a $250\times$ reduction of parameters, it remains untested if these devices can make a meaningful contribution to the global model.

\subsection{\ac{NN} Architecture Heterogeneity based on Distillation}
\label{subsec:heterogeneity_aware_distillation}
Contrary to \fedavg, \fedmd does not transfer knowledge by sharing model weights but by sharing soft labels of a public dataset. Since the aforementioned problems of \fedavg (matching of \ac{NN} features) do not apply here, the devices' capabilities can be better matched with tailored models (even disclosed from the server) as long as they share the soft label representation at the \acp{NN}' output. This allows to address hard constraints.
However, creating a suitable public dataset and distributing it to all devices may be challenging in real-world scenarios.
Additionally, training and inference on public data are computationally expensive.
Also, distributing the public dataset to all devices may induce a large communication volume, thereby increasing the communication overhead. Also, if knowledge is shared purely through soft labels, participating devices have to be stateful. Consequently, when a new device joins the system at a later stage, it still needs to train its model from scratch with the public soft labels. This is in contrast to \fedavg, where new devices simply download the latest model weights and immediately benefit from the already performed training of other devices.

\textbf{Distillation With Public Data:} Li and Wang~\cite{Li2019} introduce \fedmd (in detail shown in \cref{subsec:fedmd}), which utilizes \ac{kd} for \ac{FL}, directly addressing the heterogeneous computational capabilities of devices. They test their solution with the \emph{EMNIST/MNIST}~\cite{cohen2017emnist} and the \emph{CIFAR10/100}~\cite{krizhevsky2009learning} dataset (public/private) using 10 devices, each deploying a unique \ac{NN} architecture. They achieve a 20\% increase in accuracy on every device compared to an isolated (on-device) setting.
Chang~\etal~\cite{chang2019cronus} present \emph{Cronus}, which is similar to \fedmd and also allows for heterogeneous architectures. While in \fedmd, public and private data are trained consecutively, Cronus mixes both for local training.

\textbf{Mixture of Distillation and \fedavg:}
As distillation approaches lack behind \fedavg w.r.t. achievable accuracy, a mixture of both approaches is proposed. Lin~\etal~\cite{lin2020ensemble} present \emph{FedDF}, which moves \ac{kd} from the devices to the server, thus, removing the additional public dataset training and inference effort. FedDF, similar to \fedmd or Cronus, also allows for heterogeneous architectures, while here, the server is fully aware of the devices' architectures. Aggregation is done by averaging all devices' weights (similar to \fedavg) within groups, and building an averaged starting model for each group. Each averaged group model now acquires knowledge from averaged soft labels computed with all received devices' weights. Compared to \fedavg, FedDF shows better robustness in non-\ac{iid} data cases and allows for more local steps between rounds without degrading the performance. A disadvantage of this approach is that it requires data for distillation on the server. Shen~\etal~\cite{shen2020federated} propose \emph{FML}, where two models are deployed on each device. The first one is a custom model that best fits the devices' computational capabilities and data. The second one is a knowledge transfer model that is used with \ac{kd} to transfer knowledge between both networks in both directions. \fedavg is used to average the weights of the knowledge transfer model on a server. While they outperform \fedavg and FedProx~\cite{li2020federated} in certain experiments, this approach comes with the major downside of an additional computational burden of knowledge sharing on the device and forcing a fixed architecture for knowledge sharing on all devices.

\textbf{Single Per-Class Representations:} Transferring soft labels of a public dataset comes with a large computational burden. To address this, the transfer of single per-class representations is discussed.
Hin and Edith present \emph{FedHe}~\cite{hin2021fedhe}, which, similar to previous approaches, allows for different model architectures per device. Contrary to \emph{FedMD} or Cronus, FedHe does not use a public dataset for distillation. The devices' models only share a single per-class representation (soft label) of their output layer trained on private data with the server. On the server, the per-class representation is averaged asynchronously. The devices train on their private data with a mixture of one-hot and soft label loss. FedHe is, therefore, lightweight in communication and requires no training with public data. The authors show that FedHe outperforms FedMD in many scenarios. Similarly, Tan~\etal present \emph{FedProto}~\cite{tan2021fedproto}, where knowledge is exchanged with class prototypes instead of public soft labels. Contrary to FedHe, not the output representations of the classes are used but an internal representation (an intermediate layer output before the NN network's last layer) to allow for higher expressiveness. The models on the devices are trained with a combined loss that accounts for the one-hot encoded private data and normalizes the model by keeping representations of private samples close to the global class representations.

\textbf{Discussion:}
Presented approaches provide the most flexibility for devices to adapt their model architecture (addressing hard constraints), therefore, also achieving a finer granularity of the heterogeneity compared to using subsets. In cases where certain devices have certain ML accelerators for specific tasks or very small memory budgets, distillation-based approaches allow for specifically tailored \acp{NN} to be deployed on the devices. Still, distillation-based approaches show certain disadvantages. Firstly,  in most cases, they do not reach the same accuracy as \fedavg-based approaches. Secondly, if knowledge between server and devices is exchanged through soft labels, devices are stateful. Consequently, it is required that each device participates in a high number of rounds to have a sufficiently trained local model. Therefore, they do not scale as well as \fedavg-based techniques w.r.t. the number of participating devices and are best suited for horizontal cross-silo scenarios. Lastly, distillation adds computational overhead, limiting the applicability for purely throughput-constrained devices.

\subsection{NN Architecture Heterogeneity based on Other Techniques}
\label{subsec:heterogeneity_aware_other}
He~\etal~\cite{he2020group} present a combination of split learning and distillation in \emph{FedGKT}, where two models are employed. A lightweight feature extractor on the devices to lower the computational burden and a more complex server model. Knowledge is shared in both directions: The server receives feature maps and respective soft labels from the devices. The devices receive soft labels from the server. However, FedGKT does not allow for heterogeneous splits between device and server and provides no aggregation algorithm that supports heterogeneity. The final full model is a combination of the device and server models. The knowledge exchange is done asynchronously to avoid stragglers.
Chopra~\etal enable device heterogeneity in split learning, where the full \ac{NN} is split, and parts of the model are trained on the devices while the other part is trained on the server. They present \emph{AdaSplit}~\cite{chopra2021adasplit}, which allows for different device model sizes by varying the split position between the device and the server. While in baseline split learning, activations have to be uploaded to the server, and gradients have to be downloaded from the server, AdaSplit mitigates this by using a contrastive loss to train locally without server interaction and send activations to the server only after the local phase. The implementation allows for asynchronous transfer of gradients.

A downside of the approach is that it requires training the model partly on the server, i.e., computation of gradients by using the received activations, which is more complex than averaging and might show problems with scaling to many devices. Additionally, split learning techniques result in stateful devices, as every device's final model is a combination of device and server mode. Consequently, the presented techniques are best suited for horizontal cross-silo \ac{FL}, as each device has to participate in many rounds to reach a sufficiently trained device model.

\subsection{System Level Awareness Through Client Selection and Flexible Aggregation}
\label{subsec:heterogeneity_aware_client_selection}
\begin{figure}
    \centering
    \input{src/sync_async_system}
    \caption{\ac{FL} aggregation of three different strategies over time. The first (top) is baseline \fedavg, where constrained devices can become stragglers, slowing down the rounds. The second (middle) is \ac{FL} with system-level awareness (e.g., through client selection). The third (bottom) is asynchronous aggregation, where constrained devices can upload stale updates, hurting accuracy.}
    \label{fig:sync_async_system}
\end{figure}
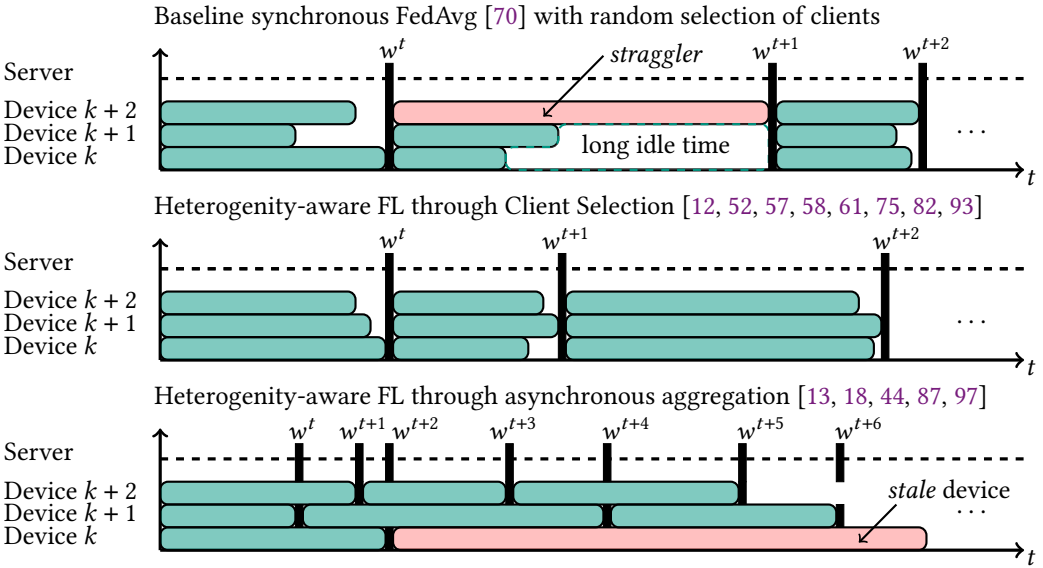
In heterogeneous \ac{FL} systems, devices with soft constraints delay the parameter aggregation since the server has to wait for the slowest device (straggler)~\cite{xu2019elfish}. To demonstrate the different behaviors over time \fedavg with stragglers is visualized in \cref{fig:sync_async_system}. Additionally, asynchronous aggregation and aggregation with system awareness are displayed. The following works account for heterogeneity by minimizing straggler time through system awareness while still maintaining synchronous rounds.

\textbf{Tier based Client Selection:} Nishio and Yonetani~\cite{nishio2019client} propose \emph{FedCS}, where they jointly consider the communication and computation resources of devices during device selection. At the beginning of each global round, the server requests the devices' current capabilities and selects a subset for the next aggregation step. The objective of this work is to achieve the highest accuracy in a limited time. Achieving this goal requires a trade-off between maximum local training time (the longer the maximum round time is, the more devices can participate) and finishing many rounds within the given time budget.
Chai~\etal\cite{Chai2020} introduce a tier-based aggregation scheme \emph{TiFL}, where devices are profiled and grouped in tiers based on the time it takes to train for one epoch. Profiling can be done either static (prior to training) or continuously. The paper discusses several tier selection schemes for training. Their experimental setup consists of five groups, where a higher tier always has the double performance of the previous tier. Training only with the fastest tier drastically reduces training time but hurts final accuracy, while uniform selection matches baseline \fedavg accuracy and reduces the training time by 50\% (\fedavg has to wait for stragglers). They propose an adaptive selection to address non-\ac{iid} data scenarios that selects tiers that have low accuracy with a higher probability, thereby achieving higher accuracy than baseline \fedavg, while significantly reducing the training time.
Reisizadeh~\etal\cite{reisizadeh2020stragglerresilient} introduce \emph{FLANP}, where the straggler problem is mitigated by utilizing the fast devices at the start. The authors assume an \ac{iid} data environment and train with fast devices until they reach an accuracy threshold. They increase the accuracy threshold and iteratively add more (slower) devices to training. This slows down training but increases accuracy since, every round, more data is available.  Using this technique, FLANP converges a lot faster compared to \fedavg when having heterogeneous devices. 

Presented techniques rely on similarities between devices, e.g., a low number of groups per round. An increase in the granularity and the scale of the heterogeneity deteriorates the gains. While an active selection of devices based on resources speeds up training in \ac{iid} scenarios, it might hurt performance in non-\ac{iid} cases, especially when resources are coupled with data distribution shifts, therefore, an environment not only enforces specific constraints on some devices but also influences the devices' data distributions.
This may, for instance, be the case in \iac{IoT} system, where some devices are deployed indoors while others are outdoors, with outdoor devices being more constrained in energy and with data distribution being variant between indoor and outdoor devices. This issue is tackled in the following techniques.

\textbf{Client Selection with Device Impact:} Lai~\etal~\cite{lai2021oort} present \emph{Oort}, a client selection framework with the aim to optimize convergence speed (w.r.t. time) by allowing for rounds with variable lengths. In difference to previous client selection techniques, Oort considers the impact a device has on the global model. Devices and round times are selected based on a utility score that includes the resources and the global model impact based on the local training loss. Li~\etal~\cite{li2022pyramidfl} introduce \emph{PyramidFL}, building on top of Oort and further improving convergence speed, mainly through a more fine-grained selection by allowing devices to train variable amounts of local epochs.

\textbf{Partial Device Updates:} Li~\etal\cite{li2020federated} introduce \emph{FedProx}, which instead of dropping stragglers completely, allows for partial contributions to the global model. This is achieved by introducing an inexactness term to the devices' updates that accounts for fewer local epochs. FedProx, therefore, allows for a variable number of local steps. Especially in cases where in baseline \fedavg many devices have to be dropped, FedProx outperforms \fedavg with respect to final accuracy. Wang~\etal~\cite{wang2020towards} present an optimization framework to tackle stragglers in a mobile device scenario. They conduct real-world measurements on smartphone processors to measure the training throughput of various devices. Contrary to other works, they also incorporate the effects of thermal throttling into the throughput model. The framework achieves optimal utilization of devices by accounting for the devices' capabilities and splitting the devices' private data into trainable subsets. The device scheduling algorithm is also designed to be aware of non-iid data.

\textbf{Efficiency Trade-Offs:}
Wang~\etal\cite{Wang2019} study how to effectively utilize available resources and obtain a convergence bound, highlighting how local device steps and global rounds contribute to convergence. Further, they propose an algorithm that finds a resource-efficient trade-off between communication and computation but does not specifically cover heterogeneity over time or across devices.
They show that in non-\ac{iid} data scenarios, their adaptive synchronous approach outperforms asynchronous settings in terms of convergence speed. Similarly, Tran~\etal~\cite{tran2019federated} study the trade-off between training time/accuracy and energy spent on training. They associate an energy cost with training throughput per time and communication bandwidth. They model energy consumption for training throughput per second and transmission of bytes per second in a wireless channel environment. A scenario with three classes of devices is studied where each device operates at certain CPU frequency levels. With that, they determine a Pareto-optimal trade-off between time and energy in FL with heterogeneous devices. Kim and Wu present \emph{AutoFL}~\cite{kim2021autofl}, a system-level approach to optimize convergence speed and energy usage by training a reinforcement learning algorithm to select an optimal subset of devices for training. The server algorithm takes several device-specific factors into account, such as the number of data classes, network bandwidth, CPU, and memory contention. The solution is evaluated in a scenario with 200 mobile devices with three different device classes (high-end, mid-end, and low-end), simulating resource contention, unstable connections, and data heterogeneity.

\textbf{Discussion:} Presented synchronous system-level approaches optimize the round time and device selection by processing the devices' resource availability information at the server mostly on a round basis. To account for soft constraints (i.e., throughput constraints) of the devices, the server has to track and monitor the devices' state and estimated capabilities. This induces overhead on the server as well as on the devices and might be unreasonable to predict in real-world scenarios. Others, like FedProx, allow partial local training, accounting for resource changes during local rounds without server knowledge.

\subsection{System-Level Awareness Through Asynchronous Aggregation}
\label{subsec:heterogeneity_aware_async}
Another way to account for different training times is by doing weight aggregation asynchronously. Thereby, each device can download the current model at any time and, depending on its resources, upload its updated weights (visualized in \cref{fig:sync_async_system}). Several papers~\cite{reisizadeh2020stragglerresilient, stich2019local, zhou18b, chen2017revisiting} hint that asynchronous aggregation achieves similar convergence speeds as synchronous schemes if the staleness of devices is within certain bounds. It has to be noted that most theoretical guarantees only cover asynchronous distributed \ac{sgd} in convex cases. Convergence speed guarantees for asynchronous \fedavg, especially the comparison with synchronous \fedavg, is ongoing research.

\textbf{Asynchronous Aggregation:} Chen~\etal\cite{chen2019asynchronous} introduce \emph{Asynchronous Online Federated Learning} (\emph{ASO}) for a setting with devices under heterogeneous resources and data quantity. They use a modified version of baseline \fedavg ($L_2$-norm regularization on the client and weight normalization on the server) to account for the devices' data quantity and reduce the effect of the devices' models drifting away from the global model. They assume devices continuously receive a new stream of training data and do not have the memory capacity to learn in batches. The authors show that their asynchronous aggregation scheme requires less time to converge in comparison to baseline \fedavg. Similarly, \emph{FedAsync} by~Xie~\etal\cite{xie2020asynchronous} uses an asynchronous aggregation scheme with a local regularization term. The staleness problem of devices is mitigated by weighting the devices' updates contribution with a time-dependent parameter. This means that the contribution of devices that take very long, thus updating an old model state of the server, is reduced. The authors show that FedAsync outperforms \fedavg in small staleness scenarios. Sprague~\etal\cite{sprague2018asynchronous} study the effects of synchronous and asynchronous \fedavg, showing that asynchronous aggregation outperforms synchronous w.r.t. convergence speed in cases where devices have different training throughput. Additionally, they study the effects of devices joining late in training, observing a disturbance in the convergence in non-\ac{iid} cases. Huba~\etal present \textit{Papaya}~\cite{huba2022papaya}, a framework for large-scale \ac{FL} that supports synchronous and asynchronous aggregation. They empirically show that in large-scale ($100$M phones) next-word-prediction tasks, asynchronous aggregation converges faster and with higher accuracy compared to synchronous \ac{FL}. In a similar experiment, synchronous \ac{FL} converges slower if the aggregation waits for stragglers or reaches lower final accuracies if stragglers are discarded.

\textbf{Hybrid Aggregation:} Extending their previous work \emph{TiFL}~\cite{Chai2020}, Chai~\etal\cite{chai2020fedat} present \emph{FedAT}, a hybrid synchronous-asynchronous approach utilizing tiers. Devices are grouped in tiers based on their performance, similar to TiFL. While devices within one tier do synchronous aggregation, tiers asynchronously update the global model. To mitigate bias towards faster tiers (especially in non-\ac{iid} cases), FedAT weights the updates of tiers, s.t. slower tiers are considered with a higher weight when updating the global model, thus allowing for equal contribution to the global model. Similar to FedProx~\cite{li2020federated}, they use a constraint term to restrict local weights to be closer to the global model. Experiments with 100 devices grouped in 5 tiers show that FedAT outperforms baseline \fedavg, as well as TiFL and pure asynchronous schemes like FedAsync~\cite{xie2020asynchronous}.

\textbf{Discussion:} Presented approaches tackle soft constraints of devices by doing aggregation asynchronously. Asynchronous approaches allow for an arbitrarily-high granularity in the heterogeneity since every device can upload its model at any time. While the straggler effect can be fully addressed that way, asynchronous aggregation suffers from stale updates. The scale of heterogeneity is therefore limited by the effect of stale updates. From current research, it can be concluded that it is unknown whether synchronous or asynchronous aggregation is ultimately preferable. Results show that, depending on the assumptions, both show advantageous properties. Further theoretical work to study convergence properties beyond convex utility cases is needed.

%% file: src/taxonomy.tex
\small
\begin{center}
	\begin{tabularx}{\textwidth}{l | c| c l | c c| c c c }
		
		\textbf{Work} & \textbf{Layer} & \multicolumn{2}{c|}{\textbf{Mechanism}} & \multicolumn{2}{c|}{\textbf{Constraints}} & \multicolumn{3}{c}{\textbf{Heterogen.}}  \\

		&  & \textbf{Async} & \textbf{Knowledge ex.} & \textbf{Hard} & \textbf{Soft} & \textbf{D} & \textbf{R} &\textbf{T}\\
        \midrule
        \multicolumn{9}{l}{\textbf{\cref{subsec:heterogeneity_aware_subsets} NN Architecture Heterogeneity based on \emph{FedAvg}}} \\

		Caldas~\etal\cite{caldas2018expanding} & Model  & - & Parameters  & \cmark & \cmark & - & - & - \\ 

		\emph{ELFISH}~Xu~\etal\cite{xu2019elfish} &Model & -  & Parameters & \cmark & \cmark & \cmark & \cmark & - \\ 

        \emph{DISTREAL} Rapp~\etal\cite{rapp2021distreal} &Model & -  & Parameters & - & \cmark & \cmark & \cmark & \cmark \\ 
		
		\emph{HeteroFL}~Diao~\etal\cite{diao2020heterofl} &Model & -  & Parameters & \cmark & \cmark & \cmark & \cmark & - \\ 

        \emph{MFL}~Yu~and Li~\cite{Yu2021Federated} &Model & -  & Parameters & \cmark & \cmark & \cmark & \cmark & - \\ 
  
		\emph{FjORD} Horv\'{a}th~\etal\cite{horvath2021fjord} &Model  & -  & Parameters &\cmark &  \cmark & \cmark & \cmark & \cmark \\ 

  		\emph{FedRolex} Alam~\etal\cite{alam2022fedrolex} &Model  & -  & Parameters &\cmark &  \cmark & \cmark & \cmark & - \\ 

  		\emph{FedorAS} Dudziak~\etal\cite{dudziak2022fedoras} &Model  & -  & Parameters &\cmark &  \cmark & \cmark & \cmark & \cmark \\ 
		
		\emph{FedHM} Yao~\etal\cite{yao2021fedhm}&Model & -  & Parameters & \cmark & \cmark & \cmark & \cmark & - \\ 

  		\emph{FLANC} Mei~\etal\cite{mei2022resource} &Model & -  & Parameters & \cmark & \cmark & \cmark & \cmark & - \\ 

  		\emph{ZeroFL} Qui~\etal\cite{qiu2022zerofl} &Model & -  & Parameters & \cmark & \cmark & - & - & - \\ 
		
		\emph{CoCoFL} Pfeiffer~\etal\cite{pfeiffer2022coco} & Model & -  & Parameters & \cmark & \cmark & \cmark & \cmark & - \\ 
        \multicolumn{9}{l}{\textbf{\cref{subsec:heterogeneity_aware_distillation} NN Architecture Heterogeneity based on Distillation}} \\
		\emph{FedMD}~Li~and~Wang \cite{Li2019}&Model & - & Soft labels (public) & \cmark & \cmark & \cmark & - & - \\ 
		
		\emph{Cronus}~Chang~\etal\cite{chang2019cronus}&Model & - & Soft labels (public) & \cmark & \cmark & \cmark & - & - \\ 
		
		\emph{FedHE} Hin~\etal~\cite{hin2021fedhe}& Model & \cmark & Soft labels (per class) & \cmark & \cmark & \cmark & \cmark & \cmark \\ 
		
		\emph{FedProto} Tan~\etal~\cite{tan2021fedproto}  & Model & - & Soft labels (per class) & \cmark & \cmark & \cmark & - & -\\  
		
		\emph{FedDF}~Lin~\etal\cite{lin2020ensemble}&Model & - & Soft labels (server) & \cmark & \cmark & \cmark & - & - \\ 
		
		\emph{FML}~Shen~\etal\cite{shen2020federated}&Model & - & Soft labels (public) & \cmark & \cmark & \cmark & - & - \\ 

        \multicolumn{9}{l}{\textbf{\cref{subsec:heterogeneity_aware_other} NN Architecture Heterogeneity based on Other Techniques}}\\

        \emph{FedGTK}~He~\etal\cite{shen2020federated}&Model & \cmark & Soft labels \& activ. & \cmark & \cmark & - & - & - \\ 

        \emph{AdaSplit}   Chopra~\etal~\cite{chopra2021adasplit} & Model & \cmark & Gradients \& activ. & \cmark & \cmark & \cmark & \cmark & \cmark \\

        \multicolumn{9}{l}{\textbf{\cref{subsec:heterogeneity_aware_client_selection} System Level Awareness Through Client Selection and Flexible Aggregation}} \\

        \emph{FedCS} Nishio and Yonetani~\cite{nishio2019client} & System & - & Parameters & - & \cmark & \cmark & - & - \\ 
		
		\emph{TiFL}~Chai~\etal\cite{Chai2020} & System & - & Parameters & - & \cmark & \cmark & \cmark & - \\ 
		
		\emph{FLANP}~Reisizadeh~\etal\cite{reisizadeh2020stragglerresilient} & System & - & Parameters & - & \cmark & \cmark & - & -\\ 

        \emph{Oort} Lai~\etal~\cite{lai2021oort} & System & - & Parameters & - & \cmark & \cmark &  \cmark & -  \\ 

        \emph{PyramidFL} Li~\etal~\cite{li2022pyramidfl} & System & - & Parameters & - & \cmark & \cmark &  \cmark & -  \\ 

        \emph{FedProx}~Li~\etal\cite{li2020federated} & System & - & Parameters & - & \cmark & \cmark & \cmark & \cmark  \\ 
  
        Wang~\etal~\cite{wang2020towards} & System & - & Parameters & - & \cmark & \cmark & \cmark & - \\ 
  
        Wang~\etal\cite{Wang2019} & System & - & Parameters & - & \cmark & - & \cmark & -  \\ 
		
        Tran~\etal~\cite{tran2019federated} & System & - & Parameters & - & \cmark & \cmark & - & -  \\ 
		
		\emph{AutoFL} Kim and Wu~\cite{kim2021autofl} & System & - & Parameters & - & \cmark & \cmark & \cmark & - \\ 
  
        \multicolumn{9}{l}{\textbf{\cref{subsec:heterogeneity_aware_async} System Level Awareness Through Asynchronous Aggregation}} \\
  
		\emph{ASO}~Chen~\etal\cite{chen2019asynchronous} & System & \cmark & Parameters & - & \cmark & \cmark & \cmark & \cmark \\ 
		
		\emph{FedAsync}~Xie~\etal\cite{xie2020asynchronous} & System & \cmark & Parameters & - & \cmark & \cmark & \cmark & \cmark \\ 
		
		Sprague~\etal\cite{sprague2018asynchronous} & System & \cmark & Parameters & - & \cmark & \cmark & \cmark & \cmark \\ 

  		\emph{Papaya} Huba~\etal\cite{huba2022papaya} & System & \cmark & Parameters & - & \cmark & \cmark & \cmark & \cmark \\ 
    
        \emph{FedAT} Chai~\etal\cite{chai2020fedat} & System & \cmark & Parameters & - & \cmark & \cmark & \cmark & \cmark \\ 
		

		\bottomrule
	\end{tabularx}
\end{center}

%% file: src/sync_async_system.tex
\begin{tikzpicture}
	\definecolor{kit}{HTML}{009682}
	
	\coordinate(start0) at (1,0) {};

	\node[yshift=0.2cm, xshift=-2.2cm, anchor=west] at  (start0) {Device $k$};
	\node[yshift=0.5cm, xshift=-2.2cm, anchor=west] at   (start0){Device $k+1$};
	\node[yshift=0.8cm, xshift=-2.2cm, anchor=west] at   (start0){Device $k+2$};
	\node[yshift=1.3cm, xshift=-2.2cm, anchor=west] at  (start0) {Server};
	
	\node[yshift=2cm, xshift=-0.2cm, anchor=west] at (start0) {Baseline synchronous \fedavg\cite{mcmahan2017communicationefficient} with random selection of clients};	
	\draw[very thick, ->] (start0) -- ++(11.5cm,0cm);
	\draw[very thick, ->] (start0) -- ++(0cm,1.6cm);
	
	\draw[thick, fill =kit, rounded corners=3pt, fill opacity=0.5] (start0) ++ (0.0cm,0.6cm) -- ++(2.6cm, 0cm) -- ++(0cm, 0.3cm) -- ++(-2.6cm,0cm) --cycle;
	\draw[thick, fill =kit, rounded corners=3pt, fill opacity=0.5] (start0) ++ (0.0cm,0.3cm) -- ++(1.8cm, 0cm) -- ++(0cm, 0.3cm) -- ++(-1.8cm,0cm) --cycle;
	\draw[thick, fill =kit, rounded corners=3pt, fill opacity=0.5] (start0) ++ (0.0cm,0.0cm) -- ++(3cm, 0cm) -- ++(0cm, 0.3cm) -- ++(-3cm,0cm) --cycle;
	
	\draw[fill=black] (start0) ++(0.0cm,0.0cm) ++(3cm, 0cm) -- ++(0cm, 1.4cm)  -- ++(0.1,0.0cm) -- ++(0cm, -1.4cm) --cycle;

	\draw[very thick, dashed] (start0) ++ (0cm,1.2cm) -- ++(11.5cm,0cm);
	
	\coordinate(start0) at (4.1,0) {};
	
	\node[yshift = 1.6cm] at (start0) {$w^{t}$};
	
	\draw[thick, fill =kit, rounded corners=3pt, fill opacity=0.5] (start0) ++ (0.0cm,0.3cm) -- ++(2.2cm, 0cm) -- ++(0cm, 0.3cm) -- ++(-2.2cm,0cm) --cycle;
	\draw[thick, fill =kit, rounded corners=3pt, fill opacity=0.5] (start0) ++ (0.0cm,0.0cm) -- ++(1.5cm, 0cm) -- ++(0cm, 0.3cm) -- ++(-1.5cm,0cm) --cycle;
	\draw[thick, fill =red, rounded corners=3pt, fill opacity=0.25] (start0) ++ (0.0cm,0.6cm) -- ++(5cm, 0cm) -- ++(0cm, 0.3cm) -- ++(-5cm,0cm) --cycle;
	
	\draw[thick, fill=white, draw=kit,dashed, rounded corners=3pt, fill opacity=0.2] (start0) ++ (1.5cm,0.0cm)  -- ++(0cm, 0.3cm) -- ++(0.7cm, 0cm) -- ++(0cm, 0.3cm) -- ++(2.8cm, 0cm) -- ++(0cm,-0.6cm) --cycle;
	
	\node[yshift = 0.3cm, xshift=3.5cm] at (start0) {long idle time};
	\node[yshift = 1.5cm, xshift=3.5cm] at (start0) (straggler) {\emph{straggler}};
	\draw[->, thick] (straggler) ++(-0.7cm,0.00cm) --++(-0.8cm,-0.8cm);

	\draw[fill=black] (start0) ++(0.0cm,0.0cm) ++(5cm, 0cm) -- ++(0cm, 1.4cm)  -- ++(0.1,0.0cm) -- ++(0cm, -1.4cm) --cycle;
	
	\coordinate(start0) at (9.2,0) {};
	
	\node[yshift = 1.6cm] at (start0) {$w^{t+1}$};
	
	\draw[thick, fill =kit, rounded corners=3pt, fill opacity=0.5] (start0) ++ (0.0cm,0.3cm) -- ++(1.6cm, 0cm) -- ++(0cm, 0.3cm) -- ++(-1.6cm,0cm) --cycle;
	\draw[thick, fill =kit, rounded corners=3pt, fill opacity=0.5] (start0) ++ (0.0cm,0.0cm) -- ++(1.8cm, 0cm) -- ++(0cm, 0.3cm) -- ++(-1.8cm,0cm) --cycle;
	\draw[thick, fill =kit, rounded corners=3pt, fill opacity=0.5] (start0) ++ (0.0cm,0.6cm) -- ++(1.9cm, 0cm) -- ++(0cm, 0.3cm) -- ++(-1.9cm,0cm) --cycle;
	
	\draw[fill=black] (start0) ++(0.0cm,0.0cm) ++(1.9cm, 0cm) -- ++(0cm, 1.4cm)  -- ++(0.1,0.0cm) -- ++(0cm, -1.4cm) --cycle;
	
	\coordinate(start0) at (11.2,0) {};
	
	\node[yshift = 1.6cm] at (start0) {$w^{t+2}$};
	
	\node[yshift = -0.1cm, xshift=1.4cm ] at (start0) {$t$};
	
	\node[yshift=+0.5cm, xshift=0.6cm] at (start0) {$\bm\ldots$};
	
	
	\coordinate(start0) at (1,-2.5) {};
	
	\node[yshift=2cm, xshift=-0.2cm, anchor=west] at (start0) {Heterogenity-aware FL through Client Selection~\cite{nishio2019client, Chai2020, reisizadeh2020stragglerresilient, li2020federated, kim2021autofl, wang2020towards, lai2021oort, li2022pyramidfl}};
	
	\draw[very thick, ->] (start0) -- ++(11.5cm,0cm);
	\draw[very thick, ->] (start0) -- ++(0cm,1.6cm);
	\draw[very thick, dashed] (start0) ++ (0cm,1.2cm) -- ++(11.5cm,0cm);

	\node[yshift=0.2cm, xshift=-2.2cm, anchor=west] at  (start0) {Device $k$};
	\node[yshift=0.5cm, xshift=-2.2cm, anchor=west] at  (start0){Device $k+1$};
	\node[yshift=0.8cm, xshift=-2.2cm, anchor=west] at  (start0){Device $k+2$};
	\node[yshift=1.3cm, xshift=-2.2cm, anchor=west] at  (start0) {Server};
	
		\draw[thick, fill =kit, rounded corners=3pt, fill opacity=0.5] (start0) ++ (0.0cm,0.6cm) -- ++(2.6cm, 0cm) -- ++(0cm, 0.3cm) -- ++(-2.6cm,0cm) --cycle;
	\draw[thick, fill =kit, rounded corners=3pt, fill opacity=0.5] (start0) ++ (0.0cm,0.3cm) -- ++(2.8cm, 0cm) -- ++(0cm, 0.3cm) -- ++(-2.8cm,0cm) --cycle;
	\draw[thick, fill =kit, rounded corners=3pt, fill opacity=0.5] (start0) ++ (0.0cm,0.0cm) -- ++(3cm, 0cm) -- ++(0cm, 0.3cm) -- ++(-3cm,0cm) --cycle;
	
	\draw[fill=black] (start0) ++(0.0cm,0.0cm) ++(3cm, 0cm) -- ++(0cm, 1.4cm)  -- ++(0.1,0.0cm) -- ++(0cm, -1.4cm) --cycle;

	\draw[very thick, dashed] (start0) ++ (0cm,1.2cm) -- ++(11.5cm,0cm);
	
	\coordinate(start0) at (4.1,-2.5) {};

	\node[yshift = 1.6cm] at (start0) {$w^{t}$};
	
	\draw[thick, fill =kit, rounded corners=3pt, fill opacity=0.5] (start0) ++ (0.0cm,0.3cm) -- ++(2.2cm, 0cm) -- ++(0cm, 0.3cm) -- ++(-2.2cm,0cm) --cycle;
	\draw[thick, fill =kit, rounded corners=3pt, fill opacity=0.5] (start0) ++ (0.0cm,0.0cm) -- ++(1.8cm, 0cm) -- ++(0cm, 0.3cm) -- ++(-1.8cm,0cm) --cycle;
	\draw[thick, fill =kit, rounded corners=3pt, fill opacity=0.5] (start0) ++ (0.0cm,0.6cm) -- ++(2cm, 0cm) -- ++(0cm, 0.3cm) -- ++(-2cm,0cm) --cycle;
	
	\draw[fill=black] (start0) ++(0.0cm,0.0cm) ++(2.2cm, 0cm) -- ++(0cm, 1.4cm)  -- ++(0.1,0.0cm) -- ++(0cm, -1.4cm) --cycle;
	
	\coordinate(start0) at (6.4,-2.5) {};
	
	\node[yshift = 1.6cm] at (start0) {$w^{t+1}$};
	
	\draw[thick, fill =kit, rounded corners=3pt, fill opacity=0.5] (start0) ++ (0.0cm,0.0cm) -- ++(4.1cm, 0cm) -- ++(0cm, 0.3cm) -- ++(-4.1cm,0cm) --cycle;
	\draw[thick, fill =kit, rounded corners=3pt, fill opacity=0.5] (start0) ++ (0.0cm,0.3cm) -- ++(4.2cm, 0cm) -- ++(0cm, 0.3cm) -- ++(-4.2cm,0cm) --cycle;
	\draw[thick, fill =kit, rounded corners=3pt, fill opacity=0.5] (start0) ++ (0.0cm,0.6cm) -- ++(3.9cm, 0cm) -- ++(0cm, 0.3cm) -- ++(-3.9cm,0cm) --cycle;
	
	\draw[fill=black] (start0) ++(0.0cm,0.0cm) ++(4.2cm, 0cm) -- ++(0cm, 1.4cm)  -- ++(0.1,0.0cm) -- ++(0cm, -1.4cm) --cycle;
	
	\coordinate(start0) at (12.3,-2.5) {};
	
	\node[yshift = 1.6cm, xshift=-1.5cm] at (start0) {$w^{t+2}$};
	\node[yshift=+0.5cm, xshift=-0.5cm] at (start0) {$\bm\ldots$};
	
	\node[yshift = -0.1cm, xshift=0.3cm ] at (start0) {$t$};

	
	\coordinate(start0) at (1,-5.0) {};
	
	\node[yshift=0.2cm, xshift=-2.2cm, anchor=west] at  (start0) {Device $k$};
	\node[yshift=0.5cm, xshift=-2.2cm, anchor=west] at  (start0){Device $k+1$};
	\node[yshift=0.8cm, xshift=-2.2cm, anchor=west] at  (start0){Device $k+2$};
	\node[yshift=1.3cm, xshift=-2.2cm, anchor=west] at  (start0) {Server};
	
	\node[yshift=2cm, xshift=-0.2cm, anchor=west] at (start0) {Heterogenity-aware FL through asynchronous aggregation~\cite{chen2019asynchronous, xie2020asynchronous, sprague2018asynchronous, chai2020fedat, huba2022papaya}};
	
	\draw[very thick, ->] (start0) -- ++(11.5cm,0cm);
	\draw[very thick, ->] (start0) -- ++(0cm,1.6cm);
	
	\draw[very thick, dashed] (start0) ++ (0cm,1.2cm) -- ++(11.5cm,0cm);
	
	\draw[thick, fill =kit, rounded corners=3pt, fill opacity=0.5] (start0) ++ (0.0cm,0.6cm) -- ++(2.6cm, 0cm) -- ++(0cm, 0.3cm) -- ++(-2.6cm,0cm) --cycle;
	\draw[thick, fill =kit, rounded corners=3pt, fill opacity=0.5] (start0) ++ (0.0cm,0.3cm) -- ++(1.8cm, 0cm) -- ++(0cm, 0.3cm) -- ++(-1.8cm,0cm) --cycle;
	\draw[thick, fill =kit, rounded corners=3pt, fill opacity=0.5] (start0) ++ (0.0cm,0.0cm) -- ++(3cm, 0cm) -- ++(0cm, 0.3cm) -- ++(-3cm,0cm) --cycle;
	
	\draw[fill=black] (start0) ++(0.0cm,0.0cm) ++(3cm, 0cm) -- ++(0cm, 0.3cm)  -- ++(0.1,0.0cm) -- ++(0cm, -0.3cm) --cycle;
	\draw[fill=black] (start0) ++(0.0cm,0.0cm) ++(3cm, 0.9cm) -- ++(0cm, 0.5cm)  -- ++(0.1,0.0cm) -- ++(0cm, -0.5cm) --cycle;
	
	\draw[fill=black] (start0) ++(0.0cm,0.0cm) ++(1.8cm, 0.3cm) -- ++(0cm, 0.3cm)  -- ++(0.1,0.0cm) -- ++(0cm, -0.3cm) --cycle;
	\draw[fill=black] (start0) ++(0.0cm,0.0cm) ++(1.8cm, 0.9cm) -- ++(0cm, 0.5cm)  -- ++(0.1,0.0cm) -- ++(0cm, -0.5cm) --cycle;
	
	\draw[fill=black] (start0) ++(0.0cm,0.0cm) ++(2.6cm, 0.6cm) -- ++(0cm, 0.3cm)  -- ++(0.1,0.0cm) -- ++(0cm, -0.3cm) --cycle;
	\draw[fill=black] (start0) ++(0.0cm,0.0cm) ++(2.6cm, 0.9cm) -- ++(0cm, 0.5cm)  -- ++(0.1,0.0cm) -- ++(0cm, -0.5cm) --cycle;
	
	\draw[thick, fill =kit, rounded corners=3pt, fill opacity=0.5] (start0) ++ (2.7cm,0.6cm) -- ++(1.9cm, 0cm) -- ++(0cm, 0.3cm) -- ++(-1.9cm,0cm) --cycle;
	\draw[thick, fill =kit, rounded corners=3pt, fill opacity=0.5] (start0) ++ (1.9cm,0.3cm) -- ++(4cm, 0cm) -- ++(0cm, 0.3cm) -- ++(-4cm,0cm) --cycle;

	\draw[fill=black] (start0) ++(0.0cm,0.0cm) ++(5.9cm, 0.3cm) -- ++(0cm, 0.3cm)  -- ++(0.1,0.0cm) -- ++(0cm, -0.3cm) --cycle;
	\draw[fill=black] (start0) ++(0.0cm,0.0cm) ++(5.9cm, 0.9cm) -- ++(0cm, 0.5cm)  -- ++(0.1,0.0cm) -- ++(0cm, -0.5cm) --cycle;
	
	\draw[fill=black] (start0) ++(0.0cm,0.0cm) ++(4.6cm, 0.6cm) -- ++(0cm, 0.3cm)  -- ++(0.1,0.0cm) -- ++(0cm, -0.3cm) --cycle;
	\draw[fill=black] (start0) ++(0.0cm,0.0cm) ++(4.6cm, 0.9cm) -- ++(0cm, 0.5cm)  -- ++(0.1,0.0cm) -- ++(0cm, -0.5cm) --cycle;
	
	\draw[thick, fill =red, rounded corners=3pt, fill opacity=0.25] (start0) ++ (3.1cm,0.0cm) -- ++(7.1cm, 0cm) -- ++(0cm, 0.3cm) -- ++(-7.1cm,0cm) --cycle;
	
	\node[xshift=10.5cm, yshift=0.8cm] at (start0) (stale) {\emph{stale} device};
	\draw[->, thick] (stale) ++(-0.7cm,-0.2cm) --++(-0.5cm,-0.5cm);

	\draw[thick, fill =kit, rounded corners=3pt, fill opacity=0.5] (start0) ++ (4.7cm,0.6cm) -- ++(3cm, 0cm) -- ++(0cm, 0.3cm) -- ++(-3cm,0cm) --cycle;
	\draw[thick, fill =kit, rounded corners=3pt, fill opacity=0.5] (start0) ++ (6.0cm,0.3cm) -- ++(3cm, 0cm) -- ++(0cm, 0.3cm) -- ++(-3cm,0cm) --cycle;
	
	\draw[fill=black] (start0) ++(0.0cm,0.0cm) ++(9.0cm, 0.3cm) -- ++(0cm, 0.3cm)  -- ++(0.1,0.0cm) -- ++(0cm, -0.3cm) --cycle;
	\draw[fill=black] (start0) ++(0.0cm,0.0cm) ++(9.0cm, 0.9cm) -- ++(0cm, 0.5cm)  -- ++(0.1,0.0cm) -- ++(0cm, -0.5cm) --cycle;
	
	\draw[fill=black] (start0) ++(0.0cm,0.0cm) ++(7.7cm, 0.6cm) -- ++(0cm, 0.3cm)  -- ++(0.1,0.0cm) -- ++(0cm, -0.3cm) --cycle;
	\draw[fill=black] (start0) ++(0.0cm,0.0cm) ++(7.7cm, 0.9cm) -- ++(0cm, 0.5cm)  -- ++(0.1,0.0cm) -- ++(0cm, -0.5cm) --cycle;
	
	\node[yshift = 1.6cm, xshift=1.9cm] at (start0) {$w^{t}$};
	\node[yshift = 1.6cm, xshift=2.7cm] at (start0) {$w^{t+1}$};
	\node[yshift = 1.6cm, xshift=3.4cm] at (start0) {$w^{t+2}$};
	
	\node[yshift = 1.6cm, xshift=4.7cm] at (start0) {$w^{t+3}$};

	\node[yshift = 1.6cm, xshift=6.2cm] at (start0) {$w^{t+4}$};
	\node[yshift = 1.6cm, xshift=8.0cm] at (start0) {$w^{t+5}$};
	\node[yshift = 1.6cm, xshift=9.3cm] at (start0) {$w^{t+6}$};
	
	\node[yshift = -0.1cm, xshift=11.6cm ] at (start0) {$t$};
	
	\node[yshift=+0.5cm, xshift=10.8cm] at (start0) {$\bm\ldots$};

\end{tikzpicture}

%% file: 5_open_problems.tex
\section{Open Problems and Future Directions}
\label{sec:open_problem}
Current techniques covered in \cref{sec:comparison} have great potential to enable \ac{FL} for applications with device heterogeneity. However, we identify several open problems that demand further research.

\textbf{Problem 1:  Maintaining effectiveness under fine-grained granularity or large-scale heterogeneity}: In most of the state-of-the-art research, tackling heterogeneity focuses on accounting for soft and/or hard constraints. The attributes scale and granularity are often neglected, are hidden behind the technique, and lack discussion in the papers. The reported scale in the resources supported by the techniques ranges from ${4\times}-{25\times}$~\cite{kim2021autofl,horvath2021fjord, mei2022resource, li2020federated, yao2021fedhm, Chai2020, shen2020federated, rapp2021distreal, pfeiffer2022coco, qiu2022zerofl} up to ${100\times}-{250\times}$~\cite{sprague2018asynchronous, diao2020heterofl}, yet it remains unclear whether training at such high scales is still effective.
Hence, while all approaches show the effectiveness of their solution in certain scenarios, it often remains unclear whether devices with low resources or stale devices can make a meaningful contribution that advances the global model. Especially in \ac{iid} settings, current state-of-the-art works do not compare themselves against trivial baselines such as dropping of devices (accepting a smaller total share of data), which is the solution that current real-world \ac{FL} applications like \emph{Google GBoard} employ. A second trivial baseline is deploying a low-complexity model for all devices~\cite{rapp2021distreal}, which can already outperform some state-of-the-art techniques.
A potential solution to maintaining effectiveness for large-scale heterogeneity with fine-grained granularity could be the interplay of system-level and NN-level approaches, as their favorable properties could complement each other. For example, an NN-architecture subset technique~\cite{diao2020heterofl, horvath2021fjord} could be complemented by system-level client selection~\cite{nishio2019client} or asynchronous aggregation~\cite{xie2020asynchronous,chen2019asynchronous,sprague2018asynchronous}  to increase granularity w.r.t. throughput. However, while system and \ac{NN} architecture level mechanisms are often orthogonal, it remains unclear how this would affect the convergence and reachable accuracy.

\textbf{Problem 2: Comparability}: Current research lacks comparability w.r.t.~the resource model. This is especially the case in techniques using subsets, where some model resources in terms of power usage~\cite{Yu2021Federated}, while others count the number of parameters~\cite{diao2020heterofl, horvath2021fjord, alam2022fedrolex, qiu2022zerofl}, the number of multiply-accumulate-operations~\cite{rapp2021distreal, horvath2021fjord}, or the required training time~\cite{pfeiffer2022coco}.
Therefore, the supported scale of heterogeneity by the techniques is not comparable.
Similarly, also the granularity of heterogeneity lacks discussion. While distillation-based approaches allow for different model architectures, only a small number of model architectures are used in the experiments (e.g., ResNet20, ResNet32~\cite{he2015deep}, and ShuffleNet~\cite{zhang2018shufflenet} in FedDF~\cite{lin2020ensemble}). The complexity differences in training these various types of networks are not further evaluated. Other approaches support $2-5$ \emph{groups} of devices~\cite{horvath2021fjord, chai2020fedat, Chai2020, tran2019federated}, which may lead to inefficient use of available resources, as discussed in \cref{subsec:computation_heterogenity_categorization}.
In general, different scales and granularities of the heterogeneity have to be taken into account to address real-world heterogeneity aspects such as varying peak performance and memory capacity, as well as resource contention.
For a broader deployment of FL solutions in real-world use cases, these aspects require further discussion.
Besides that scale of heterogeneity, also the number of devices and their share of data influences the performance of the techniques. As of now, there is no standard scenario. As a result, some approaches evaluate their techniques with over $1000$ devices, while others evaluate only a setting with $2$ participants. Benchmark scenarios to compare FL techniques have been proposed only recently~\cite{caldas1812leaf}. However, these benchmarks do not represent the resource constraints of devices as it is present in real-world applications. Additionally, available state-of-the-art \ac{FL} simulation frameworks like \emph{FLOWER}\footnote{\url{https://flower.dev}}, \emph{FedML}\footnote{\url{https://fedml.ai}}, \emph{TensorFlow Federated}\footnote{\url{https://tensorflow.org/federated}}, or \emph{OpenFL}\footnote{\url{https://github.com/intel/openfl}} do not implement device heterogeneity specifically, memory, throughput, or energy constraints. To solve this issue, firstly, a more device-representative benchmark for FL is required that more realistically models the environments of \ac{IoT}, smartphones, or sensor networks.  Secondly, heterogeneity support in popular \ac{FL} frameworks is required.

\textbf{Problem 3: Unexplored trade-offs and  non-\ac{iid} data scenarios}: The objectives of the discussed techniques are mostly accuracy or convergence speed.
Only a few consider energy efficiency in heterogeneous settings, which is crucial in many embedded or \ac{IoT} scenarios~\cite{dhar2019device} where the available energy is limited.
While utilizing all devices up to their capabilities speeds up training the most, it might not always be a very energy-efficient way to train the global model.
Limited energy leads to a trade-off between using the energy for communication or computation~\cite{Wang2019,tran2019federated}, which has not been explored in heterogeneous \ac{FL}. Besides trade-offs between communication and computation, also trade-offs between computation and memory exist, where intermediate results can either be stored in memory or dynamically recomputed when needed~\cite{kirisame2020dynamic}, which are unexplored in the context of resource-constrained \ac{FL}.

Another rather unexplored problem in \ac{FL} with computationally constrained devices is the effects of non-\ac{iid} data. A case currently not present in the literature is the case when the data distribution is non-\ac{iid} over the devices, but additionally, there is a correlation between the data and the devices' resources. Yet, a scenario like this is expected to occur in real-world FL applications~\cite{maeng2022towards}. A first example is a set of sensors with different power sources that sample environments that differ from each other. Similar examples can be found in a smartphone \ac{FL} application.  
To meet a price target for certain markets, devices with different capabilities are manufactured. Different markets can lead to differences in device usage, hence, differences in the collected data. This may lead to a non-\ac{iid} data scenario where weak devices hold a certain type of data that, due to fairness reasons, can not be excluded and has to be incorporated into the global model. This kind of non-iid scenario exacerbates the need to learn from any device available.

Further research is required to identify what the effect of these correlations is and how their effects on the global model can be mitigated to enable a fair representation of any user group in the global model.

%% file: 6_conclusion.tex
\section{Conclusion}
This survey provided an overview of \ac{FL} under computation heterogeneity among the participating devices, as it occurs in many practical scenarios.
We analyzed the computational constraints in smart devices that lead to heterogeneity and presented a categorization that groups the constraints into hard constraints and soft constraints that vary over devices, rounds, and time and can lead to heterogeneity of different scales with different granularities.
We provided a comprehensive survey on current research on \ac{FL} under heterogeneous computation constraints and how the techniques tackle the different proprieties of heterogeneity. Finally, we identify several open problems, such as the lack of comparability, problems with the solutions' effectiveness w.r.t. the heterogenities' scale and granularity, and unexplored trade-offs.